\documentclass[letterpaper, 10 pt, conference]{ieeeconf}  %

\IEEEoverridecommandlockouts                              %

\usepackage{hyperref}

\usepackage{cite}
\usepackage{amsmath,amssymb,amsfonts,mathtools,physics}
\usepackage[noend]{algorithmic}
\usepackage{algorithm}
\usepackage{graphicx}
\usepackage{subcaption}
\usepackage{tikz}
\usepackage{textcomp}
\usepackage{xcolor}
\usepackage[capitalise]{cleveref}
\usepackage{scalefnt}
\usepackage{url}

\usepackage{acro}
\def\BibTeX{{\rm B\kern-.05em{\sc i\kern-.025em b}\kern-.08em
    T\kern-.1667em\lower.7ex\hbox{E}\kern-.125emX}}

\DeclareAcronym{GOFI}{
    short   =   gofi,
    long    =   goal and occluded factor inference,
    short-format  =   \scshape,
    }
\DeclareAcronym{MCTS}{
    short   =   mcts,
    long    =   Monte Carlo tree search,
    short-format  =   \scshape,
    }
\DeclareAcronym{MAP}{
    short   =   map,
    long    =   maximum a posteriori,
    short-format  =   \scshape,
    }
\DeclareAcronym{IGP2}{
    short   =   IGP2,
    long    =   interpretable goal-based prediction and planning,
    }

\DeclareMathOperator*{\argmax}{arg\,max}
\DeclareMathOperator*{\argmin}{arg\,min}

\newcommand{\etal}{et al.\ }
\newcommand\smaller[2][0.85]{{\scalefont{#1}#2}}

\newcommand{\igp}{{\smaller{IGP2}}}

\newcommand{\defeq}{\coloneqq}
\newcommand{\ego}{\varepsilon}
\newcommand{\St}[1]{\mathcal{S}^{#1}} %
\newcommand{\st}[2]{s^{#1}_{#2}} %
\newcommand{\est}[3]{
	\ifx&#1&%
		\hat{s}^{#2}_{#3}
    \else
		\hat{s}^{#2}_{#3}(#1)
    \fi
} %
\newcommand{\ob}[2]{
	\ifx&#1&%
		o_{#2}
    \else
		o_{#2}(#1)
    \fi
} %
\newcommand{\awareset}{\mathcal{I}}
\newcommand{\Hs}[1]{
	\ifx&#1&%
		Z
    \else
		Z(#1)
    \fi
} %
\newcommand{\hs}[2]{z^{#1}_{#2}} %
\newcommand{\Act}[1]{\mathcal{A}^{#1}} %
\newcommand{\act}[2]{a^{#1}_{#2}} %
\newcommand{\Go}[1]{\mathcal{G}^{#1}} %
\newcommand{\go}[2]{g^{#1}_{#2}} %
\newcommand{\hiddenset}{\mathcal{Z}}

\begin{document}

\title{Interpretable Goal Recognition in the Presence of Occluded Factors for Autonomous Vehicles}

\author{Josiah P. Hanna$^{1,2,3}$,
Arrasy Rahman$^{1,2}$, Elliot Fosong$^{1,2}$, \\ Francisco Eiras$^{1,4}$, Mihai Dobre$^1$, John Redford$^1$, \\ Subramanian Ramamoorthy$^{1,2}$, and Stefano V. Albrecht$^{1,2}$%
\thanks{$^{1}$Five AI Ltd.\, U.K.,
         {\tt\small firstname.lastname@five.ai}}%
\thanks{$^2$School of Informatics, University of Edinburgh, U.K.}%
\thanks{$^3$Computer Sciences Department, University of Wisconsin--Madison}
\thanks{$^{4}$Department of Engineering Science, University of Oxford, U.K.}
\thanks{Correspondence to stefano.albrecht@five.ai}
}

\maketitle

\begin{abstract}
Recognising the goals or intentions of observed vehicles is a key step towards predicting the long-term future behaviour of other agents in an autonomous driving scenario.
When there are unseen obstacles or occluded vehicles in a scenario, goal recognition may be confounded by the effects of these unseen entities on the behaviour of observed vehicles. 
Existing prediction algorithms that assume rational behaviour with respect to inferred goals may fail to make accurate long-horizon predictions because they ignore the possibility that the behaviour is influenced by such unseen entities.
We introduce the Goal and Occluded Factor Inference (\acs{GOFI}) algorithm which bases inference on inverse-planning to jointly infer a probabilistic belief over goals and potential occluded factors.
We then show how these beliefs can be integrated into Monte Carlo Tree Search (\textsc{mcts}).
We demonstrate that jointly inferring goals and occluded factors leads to more accurate beliefs with respect to the true world state and allows an agent to safely navigate several scenarios where other baselines take unsafe actions leading to collisions.
\end{abstract}

\section{Introduction}\label{sec:intro}

Predicting the behaviour of other vehicles is a fundamental challenge to developing autonomous vehicles that can drive safely and effectively in the real world.
At a given moment in time, a controlled vehicle (hereafter called the \textit{ego-vehicle}) must use a limited number of observations of the current and past states of other (non-ego) vehicles to infer their future locations and velocities, and how they will respond to future ego-vehicle actions.
Accurate prediction allows the ego-vehicle to anticipate possible future collisions and plan actions that maximise the probability of safely and efficiently reaching its target location \cite{rhinehart2019precog,lee2017desire,albrecht2020integrating, chai_multipath_2019}.

Two major approaches to this problem are inverse-planning and deep-learning methods.
Inverse-planning methods assume other vehicles behave near-rationally and predict optimal paths under inferred rewards or goals \cite{albrecht2020integrating,galceran2015augmented}.
Deep-learning methods use datasets of past vehicle trajectories and learn the conditional probability of future trajectory segments given past observations and road layouts \cite{kimProbabilisticVehicleTrajectory2017,deoConvolutionalSocialPooling2018}.
While black-box deep-learning-based methods have demonstrated an impressive ability to predict future vehicle behaviour, they lack interpretability and may perform poorly if test conditions differ from how they were trained \cite{filosCanAutonomousVehicles2020,pulverPILOTEfficientPlanning2020}.
Since interpretability is a key aspect of trustworthy autonomous systems \cite{thiebes2020trustworthy}, we focus on inverse-planning prediction methods. Predicted behaviours can then be explained as rational with respect to a certain goal or cost function.

\begin{figure}[t!]
    \centering
    \begin{subfigure}{0.8\columnwidth}
    \includegraphics[width=\columnwidth]{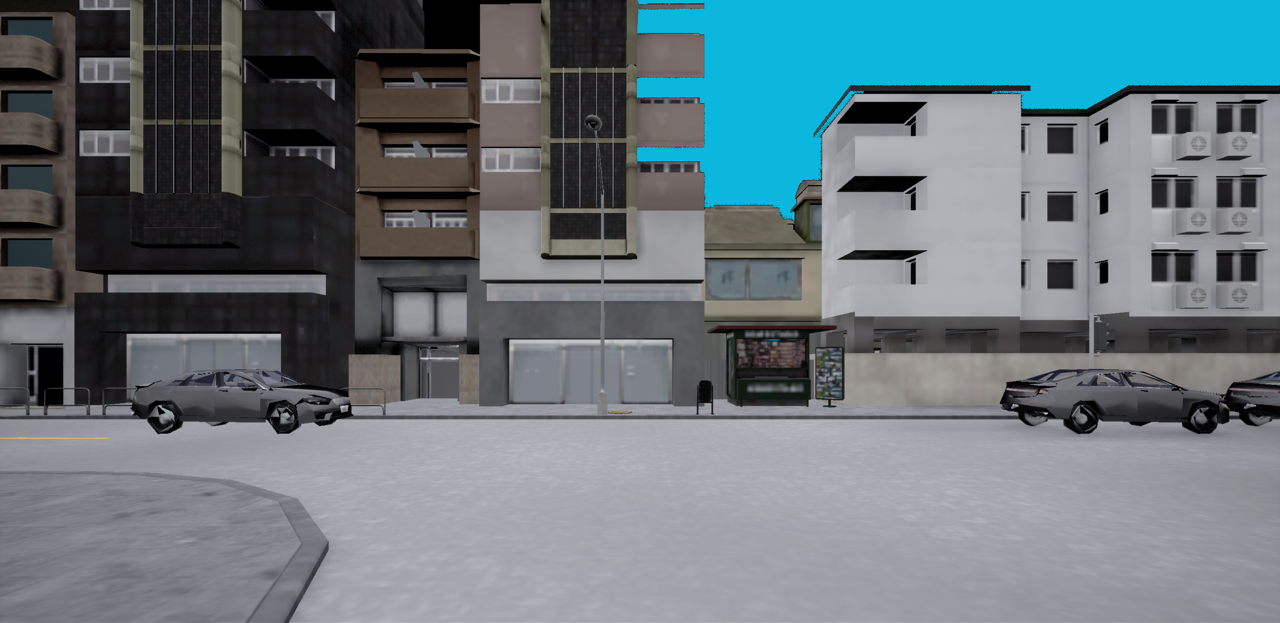}
    \caption{Ego-vehicle View}
    \end{subfigure}
    \begin{subfigure}{0.8\columnwidth}
    \includegraphics[width=\columnwidth]{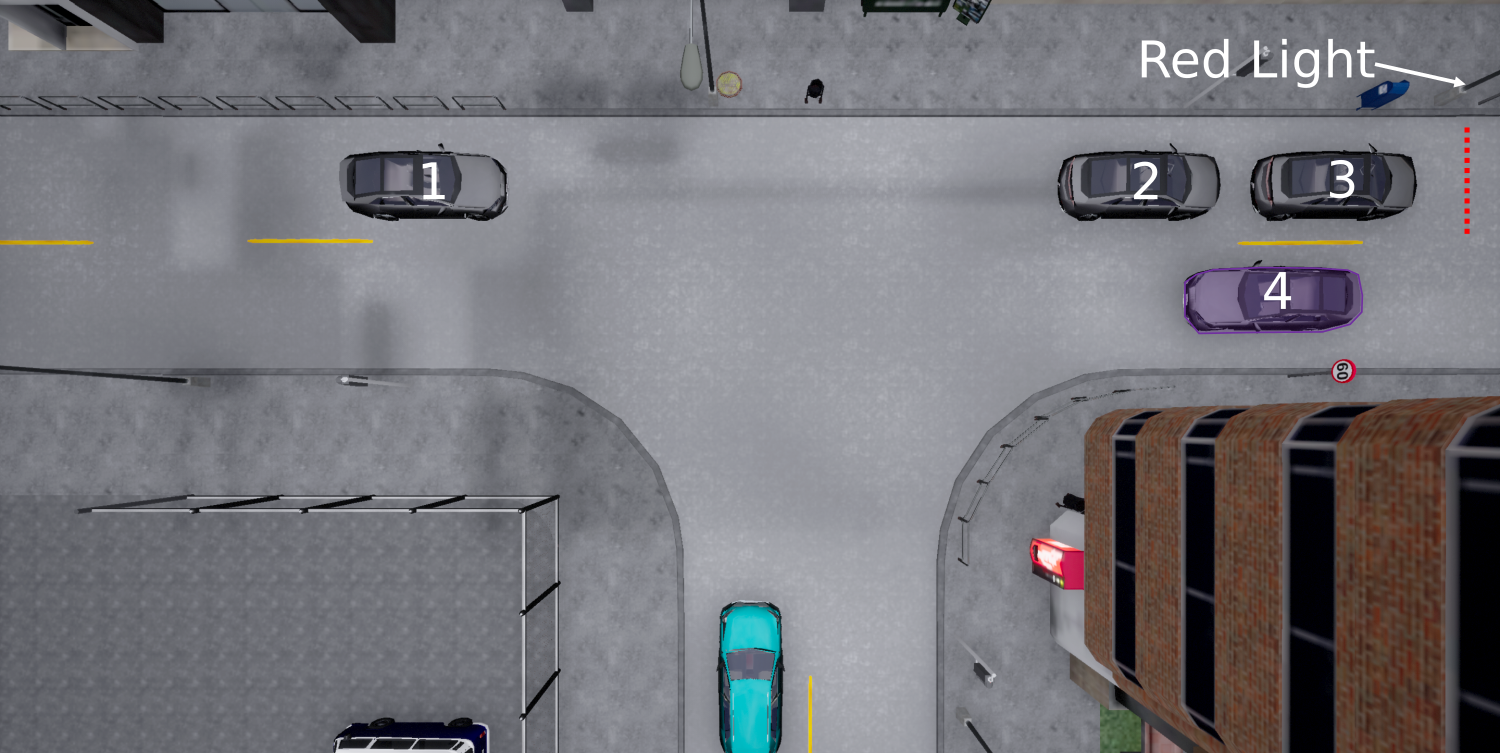}
    \caption{Bird's Eye View}
    \end{subfigure}
    \caption{Example driving scenario where the behaviour of an observed vehicle (\#1) depends on a vehicle (\#4) that the ego-vehicle (blue) cannot observe because it is occluded by a building.}
    \label{fig:occluded-parking}
\end{figure}

Recently, Albrecht \etal \cite{albrecht2020integrating} showed how goal recognition is important for long-horizon prediction.
By inferring the goals (e.g., target locations) that a non-ego vehicle is attempting to achieve and assuming the vehicle behaves approximately rationally, the space of likely future paths for that vehicle is constrained.
When goals are the main factor governing a vehicle's behaviour, goal recognition may be sufficient for explaining much of its future behaviour.
However, in a partially observable setting, the future behaviour of non-ego vehicles may depend on occluded factors that the ego-vehicle never observes.
Without considering these factors, an inverse-planning prediction method may incorrectly infer goals.
For example, consider the scenario shown in \cref{fig:occluded-parking}.
The ego-vehicle (blue) cannot see an oncoming vehicle (\#4) but observes that vehicle \#1 has stopped with a gap between it and the car in front of it.
Without knowledge of the oncoming vehicle, the ego-vehicle may conclude that \#1 has stopped to allow space for the ego-vehicle to merge in front of it.
However, an alternative explanation for the behaviour of \#1 is that it desires to turn right (and is failing to use turn signals) but is waiting for an unobserved vehicle to pass.
Thus, goal recognition is confounded by the presence of the on-coming vehicle.

The key insight behind our work is that observations of a given vehicle are a function of both that vehicle's goal and the presence or absence of potential occluded factors in the world.
Thus, we can use the past behaviour of an observed vehicle to both infer its goal and the existence of these occluded factors.
While previous work \cite{sun2019behavior,afolabi2018people} has proposed using other vehicles as ``sensors" for occluded factor inference, this prior work neglects the effects of vehicle goals or intentions on inference.

In this paper, we first describe how occluded factors  can confound prediction that relies on goal recognition.
To address this issue, we then introduce the Goal and Occluded Factor Inference (\acs{GOFI}) algorithm which resolves confounding by jointly inferring a probabilistic belief both over goals and occluded factors.
We then show how this algorithm can be integrated into Monte Carlo Tree Search (\textsc{mcts}) \cite{ks2006} to select actions for the ego-vehicle.
Finally, we empirically evaluate this algorithm on autonomous driving scenarios where reasoning about both goals and occluded factors is necessary for safe and efficient driving.
Our empirical results show that \acs{GOFI} improves inference accuracy and completes a higher percentage of trials in the evaluated scenarios.

\section{Preliminaries}
We consider a set of vehicles, $\mathcal{N}$, in which each vehicle $i \in \mathcal{N}$ has a \emph{local state} $\st{i}{t} \in \mathcal{S}$ describing its pose and velocity at timestep $t$.
We represent the ego-vehicle by index $\ego$.
The joint state at time $t$ is denoted by ${\st{}{t} \defeq (\st{1}{t}, \dots, \st{|\mathcal{N}|}{t}) \in \St{|\mathcal{N}|}}$.
The ego-vehicle $\ego$ has a potentially limited view of other vehicles in $\mathcal{N}$.
At step $t$, we assume that $\ego$ processes a sensory input sequence $\ob{}{1:t}$ to obtain a current state-estimate $\est{}{}{t} \in \St{|\awareset|}$, where
$\awareset$ is the set of all vehicles of which the ego-vehicle is aware, i.e., vehicles observed in some time-step up to and including $t$.
The processing required to obtain $\est{}{}{t}$ from $\ob{}{1:t}$ (i.e. state filtering \cite{albrecht2016causality}) is outside the scope of this paper.

We assume each vehicle $i$ has a goal, $\go{i}{}\in\Go{i}$, from a set of possible goals $\Go{i}$ and that  $\go{i}{}$ remains constant through a fixed driving scenario.
In this work, goals correspond to target locations and velocities. 
At each timestep $t$, each vehicle $i$ chooses an action $\act{i}{t} \in \Act{i}$ as a function of its goal and state-estimate.
The true state of the world then evolves as an (unknown) function of the joint action.

The ego-vehicle may be unaware of vehicles, pedestrians, or other obstacles in the environment.
We refer to such vehicles or obstacles as \textit{occluded factors}.
The possible presence or absence of occluded factors means that there are multiple world states consistent with $\est{}{}{t}$.
We define $\hiddenset \subset \St{|\mathcal{N}|}$ as the set of states consistent with $\est{}{}{t}$ where each element $\hs{}{} \in \hiddenset$ represents a possible configuration of occluded factors.
For example, in \cref{fig:occluded-parking}, the ego-vehicle can observe only cars 1--3, thus $\est{}{}{t}$ includes only their poses and velocities. 
However, this state is also consistent with the state that extends $\est{}{}{t}$ by including a vehicle with pose at the location of car 4.

In general, the size of $\hiddenset$ may be infinite, thus for tractability, we restrict ourselves to considering a finite set.
While the algorithm we introduce later is capable of reasoning over any finite subset of $\hiddenset$, in this paper we assume there are $k$ locations at which the presence of an unseen vehicle may be possible. 
In practice, suitable locations could be inferred using blindspot detection and/or a map of the static road layout.
Then, we define the set $\hiddenset \defeq \qty{\hs{}{\vb{v}} \mid {\vb{v}}\in\qty{0,1}^k }$, where $\hs{}{\vb{v}}$ corresponds to a state consistent with $\hat{s}$ but with the presence of an occluded factor at location $j$ if $v_j=1$.
We refer to elements of $\mathcal{Z}$ as \textit{occluded factor instantiations}.

Our objective is to use the observed behaviour of each ${i \in \awareset}$ to infer a probabilistic belief $p(\go{}{}, \hs{}{}| \est{}{}{1:t})$ $\forall {z\in \hiddenset},{g\in\Go{i}}$, i.e., the likelihood that vehicle $i$ has goal $\go{}{}$ and there is an occluded factor instantiation $\hs{}{}$ that influences its behaviour.
We can then use these probabilities to predict the future behaviour of vehicle $i$, infer the existence of unseen vehicles or obstacles, and plan safer actions.

\section{Related Work}

\paragraph{Goal Recognition}

Our work addresses the problem of inferring goals of autonomous vehicles as an intermediate step to predicting their future trajectory.
Goal recognition algorithms powered by deep neural networks \cite{rhinehart2019precog,lee2017desire,chai_multipath_2019,casas_intentnet_2018} have demonstrated accurate prediction by supervised-learning on large datasets.
However, these methods lack interpretability which is a challenge for their deployment in trustworthy autonomous systems \cite{koopman2019certification}.
The \igp\ algorithm of Albrecht et al.\ \cite{albrecht2020integrating} is closest to the algorithm we introduce in Section \ref{sec:inference}; we base inference on inverse-planning for interpretability and then integrate the resulting probabilistic belief into \textsc{mcts}.
However, our work differs in that we infer the presence of potential occluded factors along with goals.
As we show in our empirical analysis, a goal recognition only approach (representative of \igp) produces incorrect beliefs in the presence of occluded factors.
Outside of autonomous driving, goal (or plan) recognition has a long history in AI research \cite{albrecht2018modelling,sukthankar2014plan}.
Ramirez and Geffner \cite{ramirez2009plan} introduced one of the first goal recognition via inverse-planning algorithms.
This work inspired a number of later works which generalised the settings in which inverse-planning goal recognition could be performed \cite{sohrabi2016plan,vered2017heuristic,baker2009action,ramirez2010probabilistic}.
The most relevant to our work is the work of Ramirez and Geffner \cite{ramirez2011goal}, however, this earlier work assumes the observer (i.e., ego-vehicle) can compute the belief-state of the observed agent (i.e., non-ego vehicles).
In our setting, this assumption would amount to knowing if occluded factors are present.

\paragraph{Occluded Factors in Autonomous Driving}

Recent works have considered how occlusion affects vehicle tracking \cite{galceran2015augmented,li2018generic} and motion planning \cite{bouton2018scalable,tas2020tackling,gonzalezdebada2020oclussion}.
These problems are orthogonal to the problem of occluded factor inference from the behaviour of an observed vehicle.
The works most closely related to ours are the works of Sun \etal \cite{sun2019behavior} and Afolabi \etal \cite{afolabi2018people} who infer occluded obstacles from the actions of observed vehicles.
Sun \etal \cite{sun2019behavior} infer so-called \textit{social information} (i.e., occluded factors) with an inverse-planning approach based on a reward function learned with inverse reinforcement learning.
Afolabi \etal \cite{afolabi2018people} use human behaviour to identify an occupancy-grid representation of an environment.
Neither of these works consider goal recognition and occluded factor inference jointly.
Instead of attempting to infer the existence of occluded factors, other works attempt cautious planning when occlusion is possible.
For example, Morales \etal \cite{morales2018towards} use inverse reinforcement learning in the presence of potential occluded obstacles to learn a reward function describing how humans handle such scenarios. 
Zhang et al.\ develop a game-theoretic framework for generating behaviours that avoid collision with unobserved other vehicles \cite{zhang2021safe}.
While a cautious approach is reasonable when the full state of the world cannot be observed, our approach effectively observes possible occluded factors using the behaviour of other vehicles as a sensor.

\section{Goal and Occluded Factor Inference}\label{sec:inference}

In this section, we introduce an algorithm for inferring a probabilistic belief about a non-ego's goals and occluded factors from the behaviour of that non-ego vehicle.
We first motivate the need for joint inference of goals and occluded factors.
We then introduce an inverse-planning algorithm for conducting this inference.

\subsection{Occluded Factors Confound Goal Recognition}

Our aim is to estimate $\Pr(\go{}{}, \hs{}{}| \est{}{i}{1:t})$
for each observed vehicle $i\in\awareset$, each possible goal $\go{}{}\in\Go{i}$, and each possible occluded factor instantiation $\hs{}{}\in\hiddenset$.
Prior work has shown how $\Pr(z | \est{}{i}{1:t})$ \cite{afolabi2018people,sun2019behavior} and $\Pr(\go{}{} | \est{}{}{1:t})$ \cite{albrecht2020integrating} can be inferred in isolation for inverse-planning prediction of non-ego behaviour.
Since $\est{}{}{1:t}$ contains information about both $g^i$ and $z$, we might hope that we can simply put these prior approaches together to infer both $g$ and $z$ from the same set of past observations.
Unfortunately, since past observations, $\est{}{i}{1:t}$, depend on both $\go{i}{}$ and $z$, goals and occluded factors are dependent conditioned on past observations.
While it is reasonable to assume that goals and occluded factors are independent a priori, conditioning on $\est{}{}{1:t}$ breaks independence and necessitates joint inference, i.e., $\Pr(z|\est{}{}{1:t}) \cdot \Pr(g|\est{}{}{1:t}) \neq \Pr(g,z| \est{}{}{1:t}).$

Intuitively, an optimal trajectory for one goal may appear sub-optimal when influenced by an occluded factor that the ego-vehicle cannot observe.
The result is that the observed trajectory may provide evidence \textit{against} the non-ego's true goal instead of \textit{for} it.
This case is illustrated by the example in \cref{sec:intro} in which the non-ego vehicle pauses at the junction.
The pause is sub-optimal if the goal is to turn, however, the pause is \textit{not} sub-optimal for turning if we know there is an occluded vehicle.

\subsection{An Algorithm for Goal and Occluded Factor Inference}

To address this challenge, we introduce the \textbf{G}oal and \textbf{O}ccluded \textbf{F}actor \textbf{I}nference (\acs{GOFI}) algorithm.
\acs{GOFI} outputs a probabilistic belief over possible occluded factor states, $\Pr(\hs{}{} | \est{}{i}{1:t})$, and goals of the non-ego, $\Pr(\go{}{} | \hs{}{}, \est{}{i}{1:t})$. %
These beliefs can then be used downstream for ego-vehicle planning, as we show in \cref{sec:planning}.

\acs{GOFI} is based on the rational inverse-planning \igp\ algorithm introduced by Albrecht \etal \cite{albrecht2020integrating}.
However, since \igp\ does not model occluded factors, it may determine that a vehicle is acting irrationally with respect to one goal even if it is perfectly rational once occluded factors are taken into account.
Our main algorithmic contribution is to extend this algorithm to handle such occluded factors in goal recognition and to infer occluded factors along with goals.

Rational inverse-planning computes the optimal plan for a non-ego vehicle to achieve a potential goal given a potential instantiation of occluded factors.
The joint probability of that goal and occluded factor instantiation is computed based on the difference between the non-ego's observed behaviour and this optimal plan.
Specifically, let $\st{\star i}{1:T}(\go{}{},z)$ denote the optimal trajectory for vehicle $i$ from $\st{i}{1}$ to $\go{}{}$ and let $c^\star(g,z) \in \mathbb{R}^+_0$ be its cost under a pre-defined cost function.
Let $\st{+i}{1:T}$ be the concatenation of the observed trajectory $\est{}{i}{1:t}$ and the optimal trajectory from $\st{i}{t}$ to $\go{}{}$; let $c^{+}(\go{}{},\hs{}{}) \in \mathbb{R}^+_0$ be its cost.
\acs{GOFI} begins by computing the unnormalised likelihood of the observed trajectory for an observed non-ego vehicle $i$ conditioned on a particular $\go{}{}$ and $\hs{}{}$ instantiation.
Following Baker \etal \cite{baker2009action} and Ramirez and Geffner \cite{ramirez2010probabilistic}, we parameterise the likelihood as a Boltzmann distribution with temperature $\beta^{-1}$:
\begin{equation}
 L(\est{}{i}{1:t}|g,z) \defeq \exp(\beta( c^\star(\go{}{},\hs{}{}) - c^+(\go{}{},\hs{}{}))).   
\end{equation}
This computation is repeated for each potential goal and occluded factor instantiation.
From the computed likelihoods, the joint probability of any occluded factor state and goal can be computed with Bayes' rule as:
\begin{equation}
      \Pr(\go{}{}, \hs{}{} | \est{}{i}{1:t}) \propto L(\est{}{i}{1:t} | \go{}{}, \hs{}{}) p(\go{}{})p(\hs{}{}).  
\end{equation}
We then obtain $\Pr(\hs{}{}|\est{}{i}{1:t})$ and $\Pr(\go{}{}|\est{}{i}{1:t},\hs{}{})$ via marginalisation and a further application of Bayes rule.
Pseudocode for \acs{GOFI} is given in \cref{alg:goalrec}.

\begin{algorithm}
	\textbf{Input:} vehicle $i$, state estimates $\est{}{}{1:t}$, possible goals $\Go{i}$, set of occluded factor $\hiddenset$ \\
	\textbf{Returns:} occluded factor probabilities $\Pr(\hs{}{} | \est{}{i}{1:t})$ and goal probabilities $p(\go{i}{} | \est{}{i}{1:t}, \hs{}{})$
	
	\begin{algorithmic}[1]

		\algsetup{linenodelimiter=: }
		\STATE Set prior probabilities $p(\go{i}{})$, $p(\hs{}{})$ (e.g. uniform)
		\FORALL{$\hs{}{} \in \hiddenset$}
    		\FORALL{$g^i \in \Go{i}$}
    		    \STATE $\st{\star i}{1:T}\gets$ \textsc{PlanOptimal}($\est{}{i}{1},\go{i}{},\hs{}{}$)   %
    		    \STATE $c^\star \gets$ $\mathtt{cost}(\st{\star i}{1:T}, \hs{}{})$
    		    \STATE $\st{+i}{t+1:T} \gets$ \textsc{PlanOptimal}($\est{}{i}{t},\go{i}{},\hs{}{}$)   %
    		    \STATE  $\st{+i}{1:t} \gets \est{}{i}{1:t}$
    		    \STATE $c^{+} \gets$ $\mathtt{cost}(\st{+i}{1:T}, z^i)$
    		    \STATE $L(\est{}{i}{1:t} | \go{i}{}, \hs{}{}) \gets \exp( \beta (c^\star - c^+) )$
    		\ENDFOR
    	\ENDFOR
    	\STATE $\Pr(\hs{}{} | \est{}{i}{1:t}) \propto \sum_{\go{}{}} L(\est{}{i}{1:t} | \go{}{}, \hs{}{}) \, p(\go{}{})p(\hs{}{})$
    	\STATE $\Pr(\go{i}{} | \est{}{i}{1:t}, \hs{}{}) \gets  L(\est{}{i}{1:t} | \go{i}{}, \hs{}{}) \,  p(\go{i}{}) p(\hs{}{})/ p(\hs{}{} | \est{}{i}{1:t})$
		\STATE Return $\Pr(\hs{}{} | \est{}{i}{1:t})$, $\Pr(\go{i}{} | \est{}{i}{1:t}, \hs{}{})$
	\end{algorithmic}
    	
	\caption{Goal and Occluded Factor Inference (\acs{GOFI})}
	\label{alg:goalrec}
\end{algorithm}

Implementing \acs{GOFI} requires knowledge of the set of possible goals and occluded factors.
The set of possible goals can be computed heuristically, e.g., using the static road layout and positing possible goals at landmarks such as the end of visible lanes.
The set of possible occluded factors could be obtained by determining where blindspots are from a perception module and a map of the static road layout.
The main limitation of \acs{GOFI} is that it involves computing $2 |\Go{}| |\hiddenset|$ plans which may be computationally demanding in heavily occluded scenarios such as in dense traffic.
Fortunately, the structure of \acs{GOFI} lends itself to a parallel implementation and thus the run time can (in principle) be reduced to close to the time for the longest running call to \textsc{PlanOptimal} and subsequent cost function computation.

\acs{GOFI} also makes use of a prior distribution over elements of each of these sets (\cref{alg:goalrec}, Line 1).
These prior distributions can be used to encode prior knowledge about goals and occluded factors (e.g., historical traffic data suggesting that in the evening most vehicles are turning into a nearby residential area) or incorporate information from perception.
Signals from other vehicles can also be incorporated into priors: a vehicle which is signalling to turn is likely to have a goal that requires the turn (though it is \textit{not} certain to turn, as the signal may be accidental).
In the absence of such knowledge these priors can default to the uniform distribution.

A final consideration for \acs{GOFI} is how to merge the ego's beliefs when observing multiple vehicles.
Goals for each vehicle can be inferred independently from each vehicle \cite{albrecht2020integrating}, however, the set of occluded factors is shared across vehicles.
Thus, runs of \acs{GOFI} for each vehicle may produce different beliefs $\Pr(z | \est{}{}{1:t})$.
Belief merging across multiple non-ego vehicles can be handled by using the posterior distribution over occluded factors given the observed trajectory of one vehicle as the prior distribution when processing the observed trajectory of a second vehicle.

\section{Integrating Inference into Planning}\label{sec:planning}

Ultimately, we seek to infer vehicle goals and occluded factor instantiations to improve ego-vehicle action selection.
In this section we show how an inferred belief about goals and occluded factors can be integrated into \textsc{mcts} \cite{ks2006} action selection.
\textsc{mcts} builds a search tree of possible ego-vehicle action sequences by running a fixed number of iterations, exploring possible action sequences.
At the start of each iteration, we sample the occluded factor instantiation and goal instantiations for each observed non-ego vehicle from the belief output by \acs{GOFI}.\footnote{When considering multiple non-ego vehicles, the occluded factor instantiation should be sampled from the merged belief across vehicles as described in the previous section.}
The sampled instantiations are then treated as the ground-truth occluded factor instantiation and goal assignments for that iteration of \textsc{mcts}.
Using sampled beliefs as ground-truth is known as \textit{determinisation} in the planning under partial observability literature \cite{whitehouse2011determinization}.
Full pseudo-code for this procedure is given in Appendix \ref{appendix:mcts}.

\textsc{mcts} can be used with either finite or real-valued action-spaces \cite{kurzer2020parallelization,lee2020monte}.
To facilitate long-horizon planning, we follow Albrecht et al.\ \cite{albrecht2020integrating} and plan using \textit{macro-actions} where a macro-action is a sequence of pre-defined maneuvers such as turn, change-lane, or follow-lane.
An example of a macro-action is ``follow lane to exit" which entails a follow-lane maneuver followed by a turn.
Maneuvers define trajectories consisting of a reference path and target speeds along the path, which are mapped to steering and acceleration controls via \textsc{pid} control or adaptive cruise control.

The key requirement for this planning process is to provide adequate behaviour models for both observed and non-ego vehicles, and inferred vehicles or obstacles corresponding to occluded factors.
We assume observed vehicles follow the optimal trajectory to the sampled goal.
We assume that inferred vehicles corresponding to occluded factors follow simple behaviours such as constant velocity lane-following or a stationary pose for inferred static obstacles.
This model is convenient since these optimal trajectories are already computed by \acs{GOFI}.
To reduce computation time, we pre-sample trajectories for each vehicle (both observed and inferred) at the beginning of each \ac{MCTS} iteration.
The ego-vehicle then plans to find maneuvers that avoid collisions with these trajectories.

\section{Empirical Evaluation}

We next design and conduct an empirical analysis to answer the questions: (1) Does joint inference of goals and occluded factors increase inference accuracy?; (2) Does \acs{GOFI} increase the safety of \ac{MCTS} planning in the presence of occluded factors?

\subsection{Empirical Set-up}
\begin{figure}[t]
    \centering
    \begin{subfigure}{0.49\linewidth}
    \includegraphics[width=\linewidth]{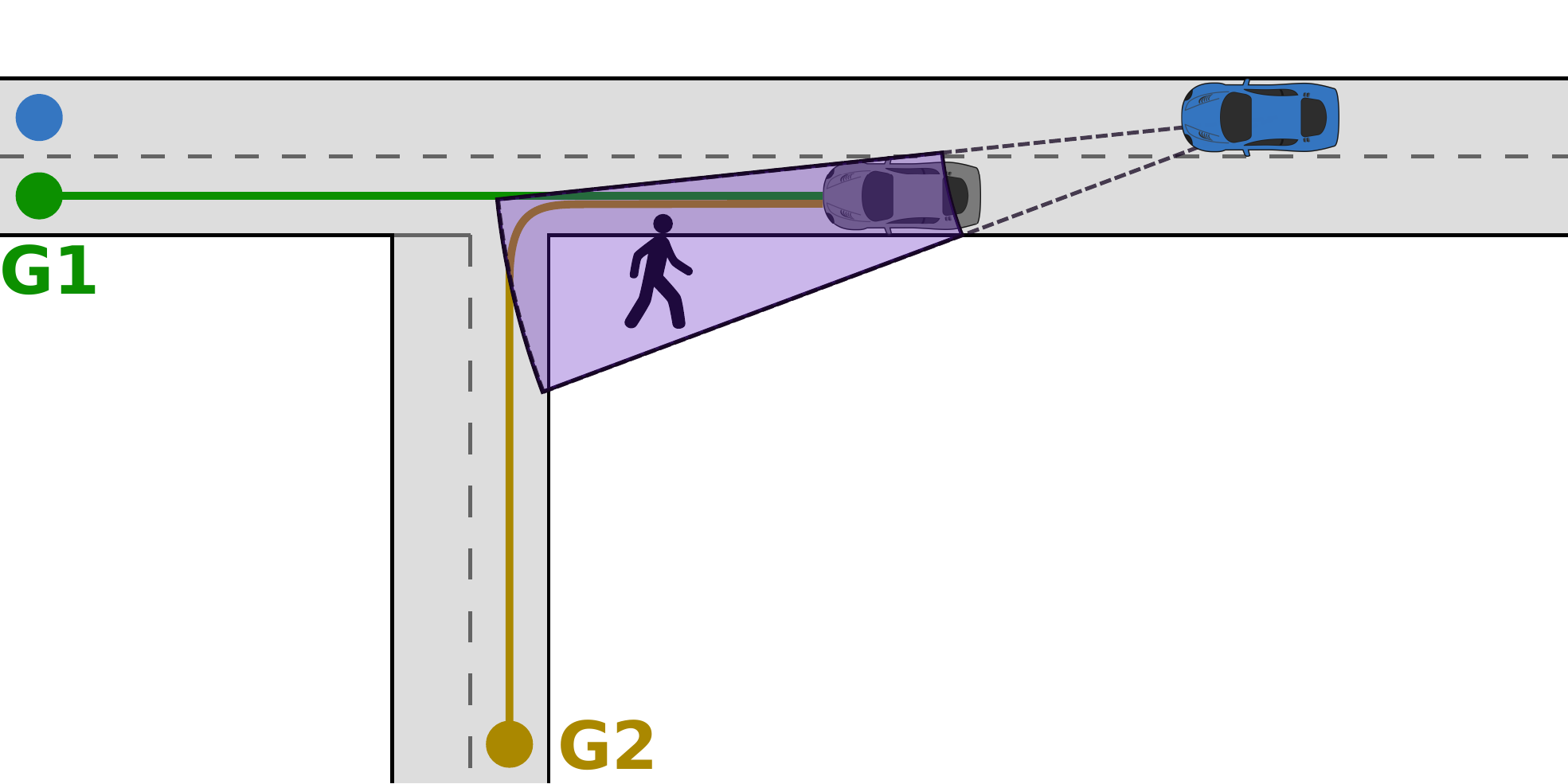}
    \caption{Scenario 1}
    \label{fig:s1}
    \end{subfigure}
    \begin{subfigure}{0.49\linewidth}
    \includegraphics[width=\linewidth]{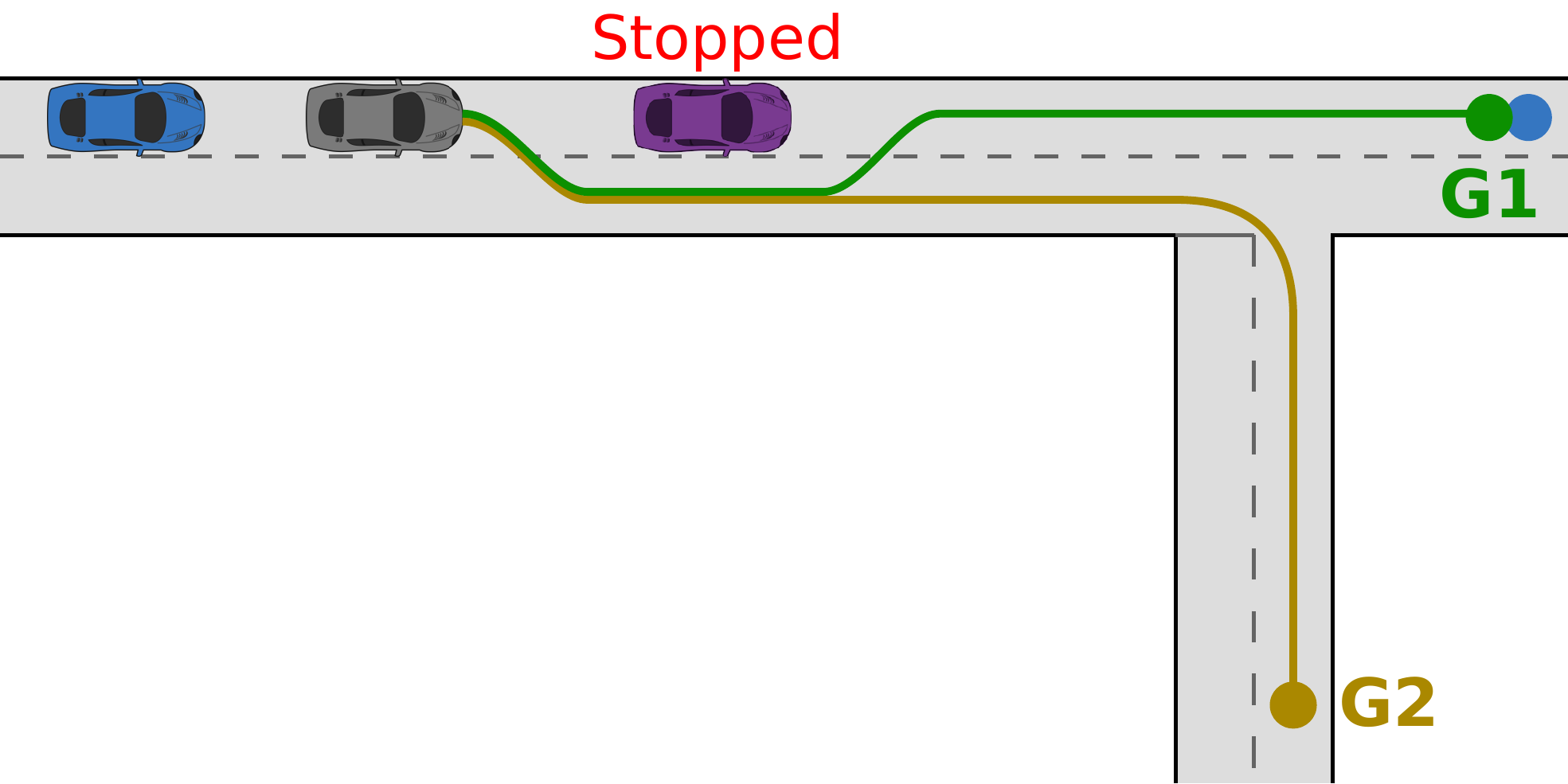}
    \caption{Scenario 2}
    \label{fig:s2}
    \end{subfigure}
    
    \begin{subfigure}{0.49\linewidth}
    \includegraphics[width=\linewidth]{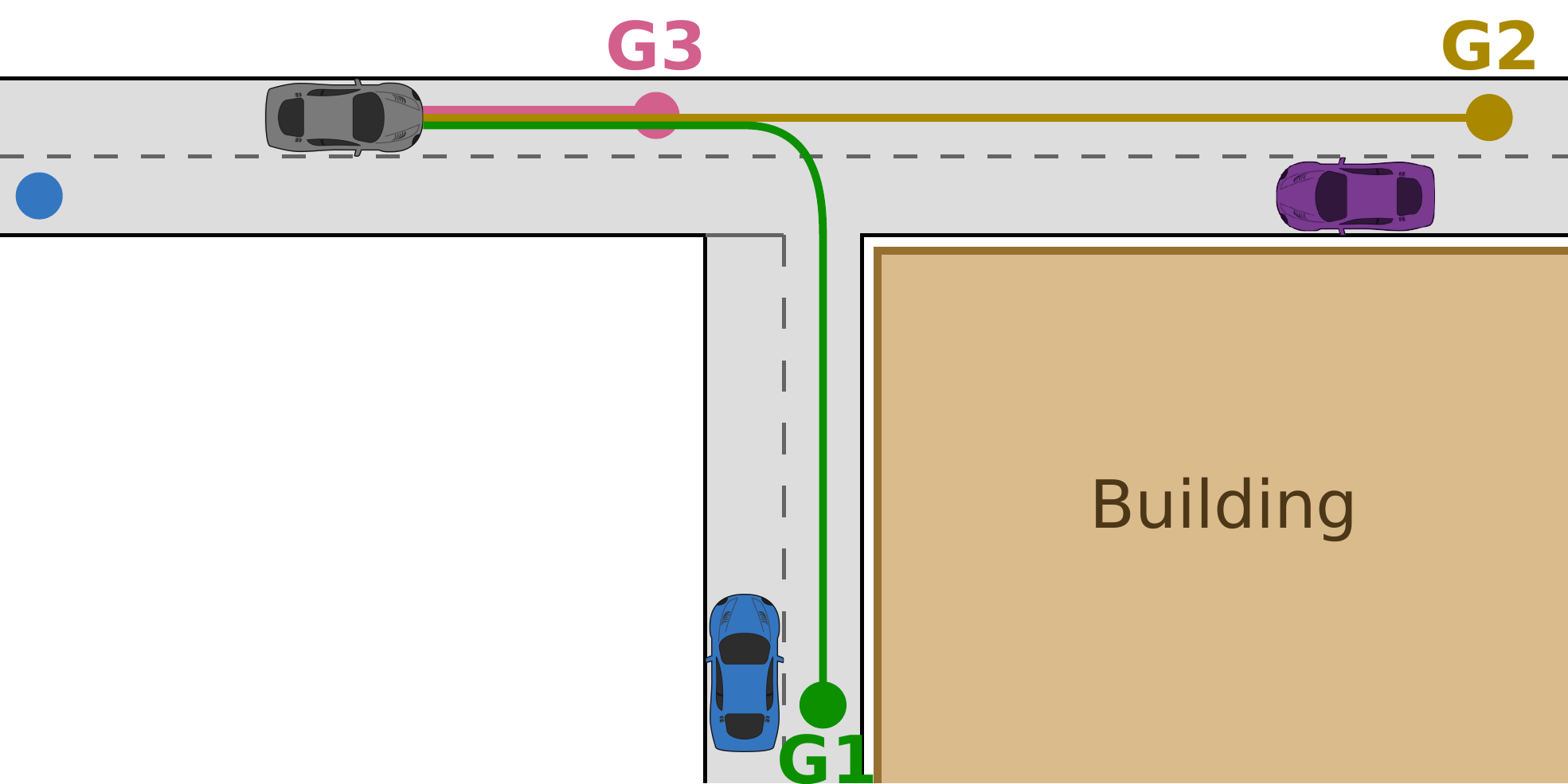}
    \caption{Scenario 3}
    \label{fig:s3}
    \end{subfigure}
    \begin{subfigure}{0.49\linewidth}
    \includegraphics[width=\linewidth]{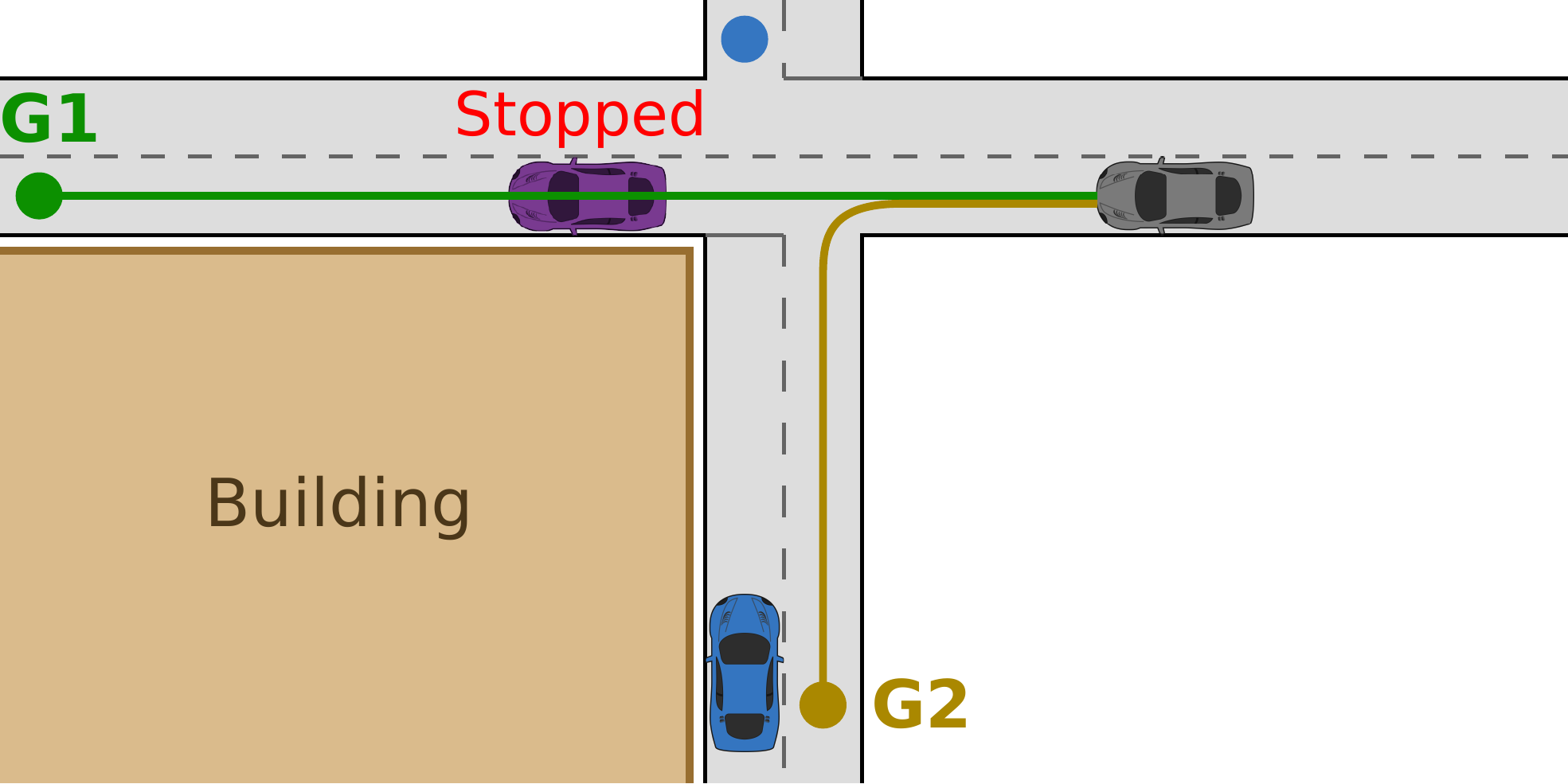}
    \caption{Scenario 4}
    \label{fig:s4}
    \end{subfigure}    
    \caption{
        \textbf{Evaluation Scenarios}.
        In all scenarios the ego-vehicle is shown in blue; non-ego, observed vehicles are grey; and occluded factor vehicles are purple.
        The ego-vehicle's goal is represented by a blue circle. Possible goals for the non-ego observed vehicle are shown by yellow/green circles, with optimal paths to each goal. In each case, the true goal of the non-ego observed vehicle is G1.
        }
    \label{fig:scenarios}
\end{figure}

To answer our empirical questions, we conduct experiments in fixed-frame driving scenarios (shown in \cref{fig:scenarios}).\footnote{Scenario videos can be found at: \url{https://www.five.ai/gofi}}
These scenarios were identified as realistic settings where the presence of occluded factors confound goal recognition and where recognition of occluded factors is necessary to avoid collisions.
\acs{GOFI} provides a general framework for joint inference and is instantiated with a particular method for \textsc{PlanOptimal}.
In our experimental implementation, we implement \textsc{PlanOptimal} with A* search over a finite set of pre-defined maneuvers.
Implementation details for A* follow Albrecht \etal \cite{albrecht2020integrating}.

\textbf{Baseline Methods:}
We use the following methods as prediction and planning baselines.
    1) \textbf{GR-Only:} performs inverse-planning goal recognition based on Albrecht \etal \cite{albrecht2020integrating} without modelling occluded factors. This baseline is implemented by running \cref{alg:goalrec} with $\hiddenset$ only consisting of the state corresponding to no occluded factors.
    2) \textbf{OF-Oracle:} the same as GR-Only except $\hiddenset$ only consists of the state corresponding to occluded factors present. This approach serves as an upper bound for inferring the true goals when occluded factors are present (as they are in our main experiments).
    3) \textbf{Goal-Oracle:} performs occluded factor inference assuming the observed vehicle's goal is known, i.e., \cref{alg:goalrec} with $\Go{i}{} = \{g^i\}$. This approach is an upper bound for inferring occluded factors.
    4) \textbf{MAP:} uses a maximum a posteriori combination of \acs{GOFI} and \ac{MCTS}. \acs{GOFI} is used for goal and occluded factor inference but only the most likely occluded factor instantiation and inferred goal are considered for predicting the future trajectory of non-ego vehicles during \ac{MCTS}.
In all methods, we use a uniform prior on the possible goals and a prior of $0.1$ on the occluded factor instantiation with an occluded vehicle present.
We use a lower prior on an occluded vehicle's presence since, in the absence of other evidence, we wish to avoid assuming existence of an occluded vehicle which could lead to overly cautious driving.

\textbf{Evaluation Scenarios:}
In \textbf{Scenario 1}, the non-ego vehicle blocks the ego's view of a pedestrian crossing the road before a junction. The non-ego vehicle slows down to allow the pedestrian to cross into the ego's lane. This behaviour could also indicate preparation to turn.
In \textbf{Scenario 2}, the ego-vehicle and observed non-ego travel to the end of their initial lane. The observed non-ego changes lane to avoid a stopped vehicle that the ego-vehicle
cannot observe. It then switches back to the left lane after overtaking the stopped vehicle.
In \textbf{Scenario 3}, a building obscures the ego’s view of an oncoming vehicle while the observed non-ego stops and waits.  The wait could indicate that the non-ego's goal is to turn.
In \textbf{Scenario 4}, the ego-vehicle cannot observe a stopped car around the corner that causes the non-ego vehicle to slow down as it approaches the junction. The slow down could also be indicative of a vehicle preparing to turn (without signalling).
Each scenario considers a single occluded factor and a single non-ego vehicle which has two possible goals.
Inaccurate inference may result in a collision with the occluded factor (Scenarios 1-3) or the visible non-ego vehicle (Scenario 4).
The initial position and velocity of each vehicle is randomised (except for the static occluded vehicles in Scenarios 2 and 4).

Since our evaluation focuses on behaviour-based inference of occluded factors, in our main experiments we first test a setting where the ego-vehicle is always blind to occluded vehicles and pedestrians -- they can only be inferred from the behaviour of the observed non-ego vehicle.
In some of our scenarios, the ego-vehicle would eventually be able to see the occluded factor.
Thus, in \cref{sec:empirical:mcts}, we also evaluate safe driving in Scenarios 1, 2 and 3 when the ego-vehicle can eventually observe the occluded factors.
These experiments show how behaviour-based inference of occluded factors complements perception and leads to safer driving.

\subsection{Inference Evaluation}

\begin{figure*}
\centering
    \begin{subfigure}{0.24\linewidth}
        \includegraphics[width=\linewidth,clip=true,trim=2cm 1cm 2cm 1cm]{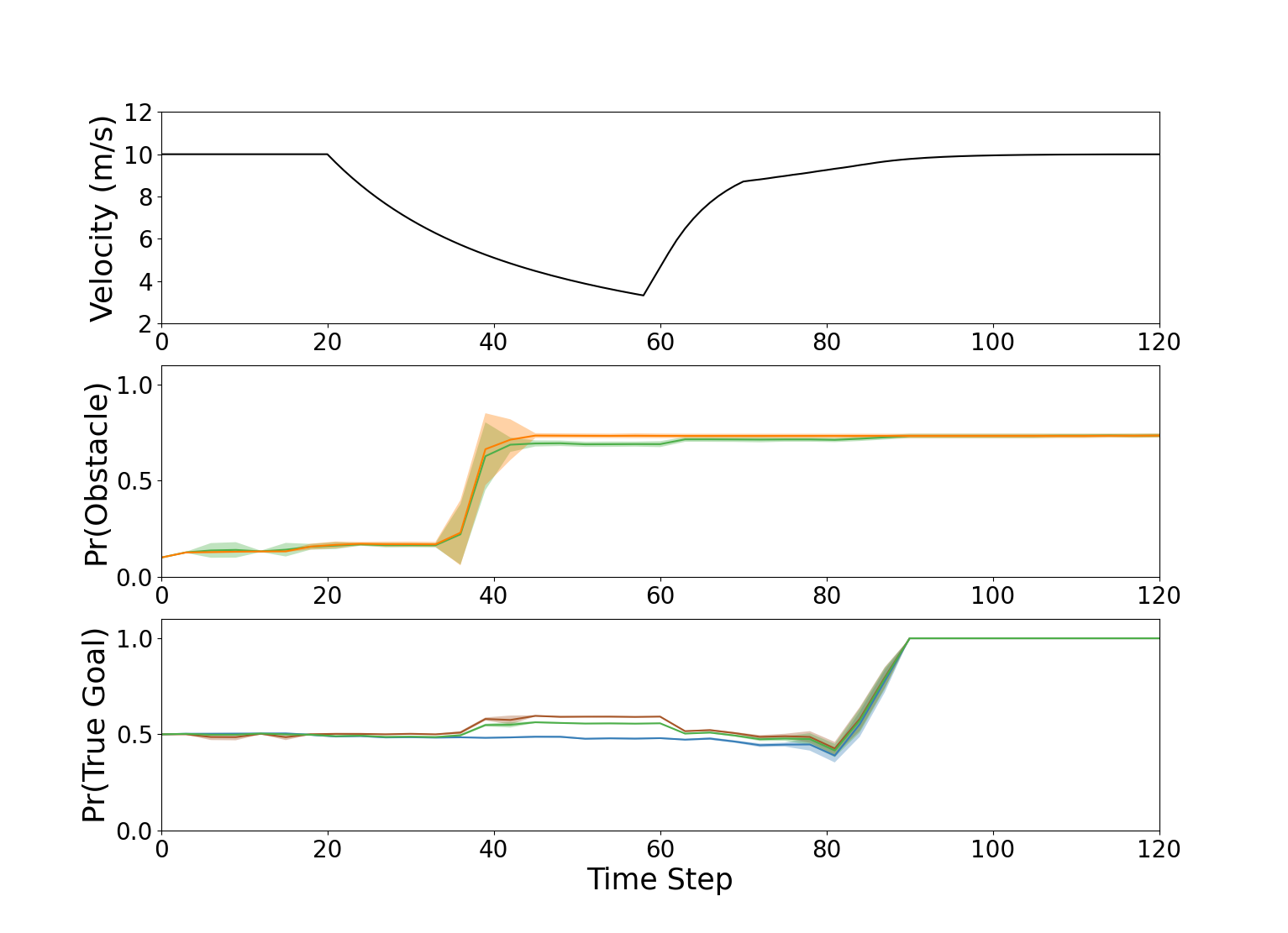}
    \caption{Scenario 1}
    \label{fig:s1:probs}
    \end{subfigure}
    \begin{subfigure}{0.24\linewidth}
        \includegraphics[width=\linewidth,clip=true,trim=25mm 1cm 30mm 1cm]{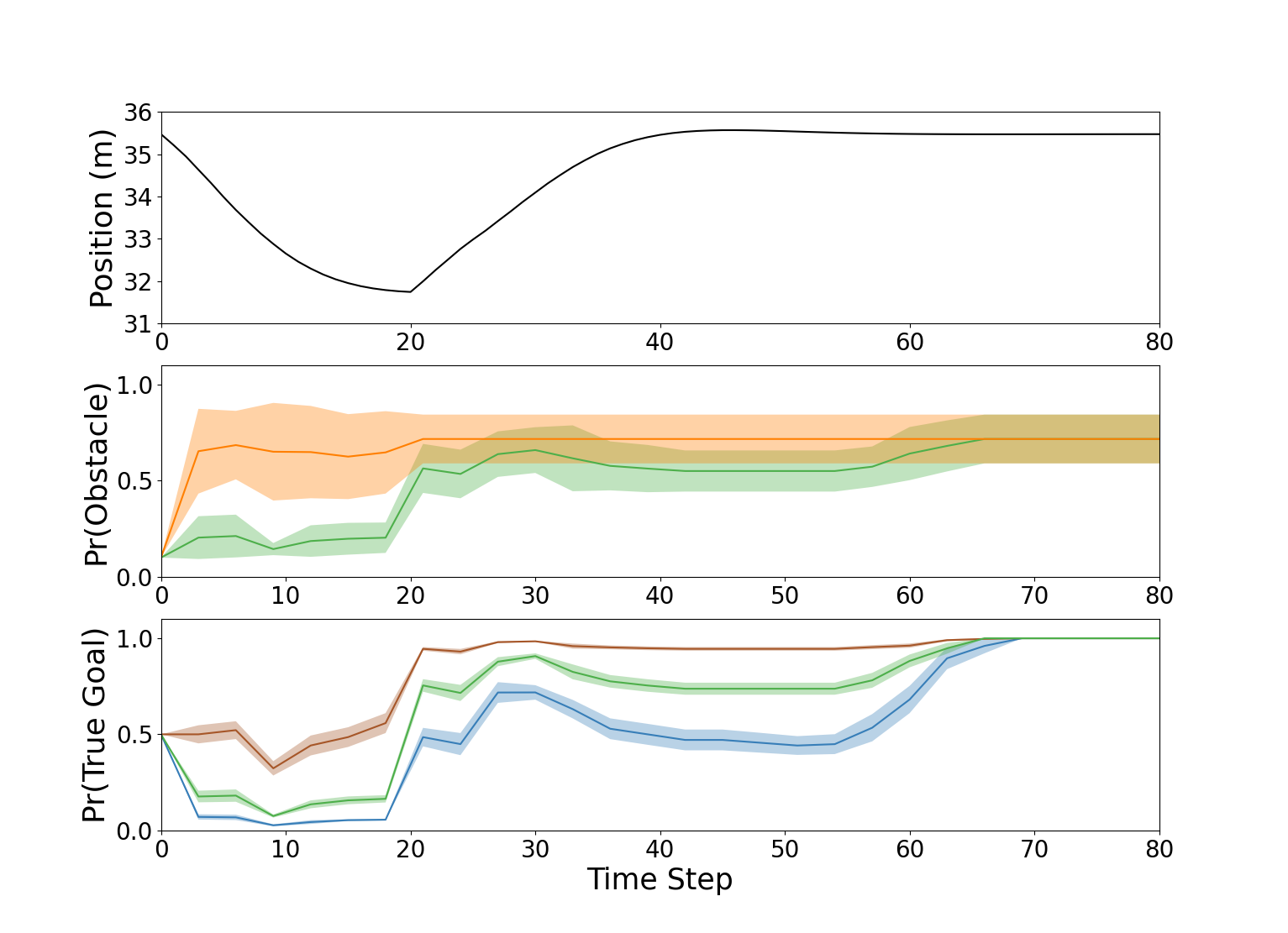}
    \caption{Scenario 2}
    \label{fig:s2:probs}
    \end{subfigure}
    \begin{subfigure}{0.24\linewidth}
        \includegraphics[width=\linewidth,clip=true,trim=25mm 1cm 30mm 1cm]{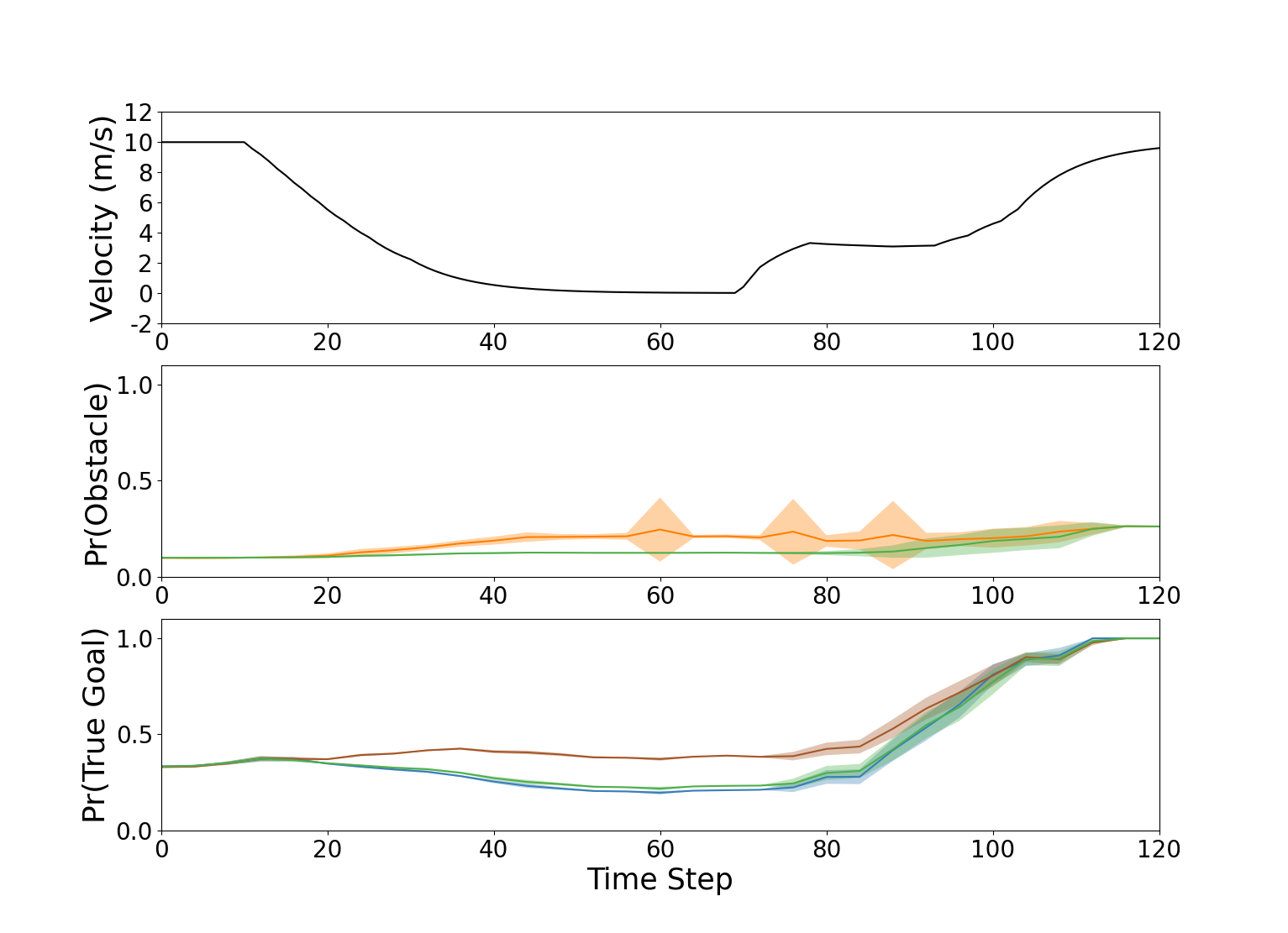}
    \caption{Scenario 3}
    \label{fig:s3:probs}
    \end{subfigure}
    \begin{subfigure}{0.24\linewidth}
        \includegraphics[width=\linewidth,clip=true,trim=25mm 1cm 30mm 1cm]{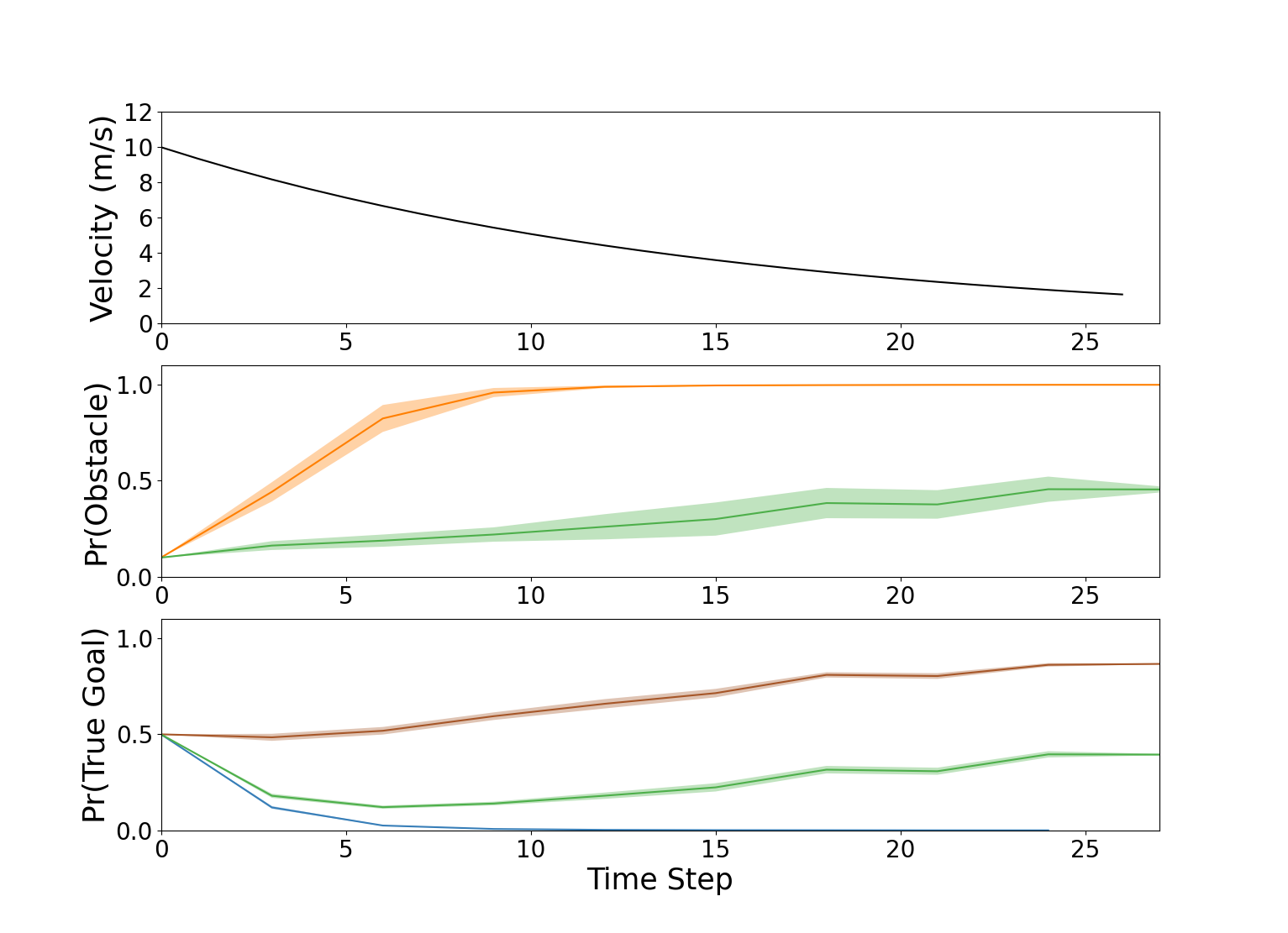}
    \caption{Scenario 4}
    \label{fig:s4:probs}
    \end{subfigure}
    \centering
    \begin{subfigure}{\linewidth}
        \includegraphics[width=\linewidth,clip=true,trim=0cm 0cm 0cm 0cm]{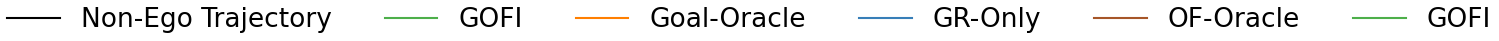}
    \end{subfigure}
    \caption{
        \textbf{Scenario Results.}
        \textit{Top}: Trajectory of the observed non-ego vehicle. In \cref{fig:s2:probs}, we plot the vertical position of the non-ego over time since it better describes the non-ego's behavior than velocity.
        \textit{Middle}: Posterior probability for the belief in the presence of an occluded obstacle over time (higher is better).
        \textit{Bottom}: Posterior probability for belief in the true goal over time.
        Beliefs are averaged over 20 repetitions (shaded area represents one standard error).
        }
    \label{fig:probs}
\end{figure*}

To answer our first empirical question, we run \acs{GOFI} and baselines to infer goals and occluded factors for the non-ego vehicle in each scenario.
We compare \acs{GOFI} and baselines in terms of their posterior probability on the true occluded factor and true goal for the non-ego vehicle (higher is better).
These posterior probabilities are shown in \cref{fig:probs}.

From the middle plots in each subfigure of \cref{fig:probs}, we can observe that \acs{GOFI} is always able to increase its posterior belief in the presence of an occluded vehicle as it observes more of the non-ego's behaviour.
The Goal-Oracle method is given the correct goal of the observed non-ego and thus only has to consider occluded factors. This makes its posterior an upper bound for \acs{GOFI}. However, in some scenarios we observe that \acs{GOFI} is near this upper bound.

The bottom plots in each subfigure of \cref{fig:probs} give the posterior belief in the correct non-ego goal.
In these plots, OF-Oracle is aware of the presence of the occluded factor and thus serves as an upper bound on obtainable posteriors. We note that in some scenarios there is sufficient ambiguity such that, even with this knowledge, the true goal is hard to infer.
On the other hand, GR-Only does \textit{not} model the possibility of occluded factors and thus obtains the lowest posterior beliefs on the true goal. In one case, GR-Only converges to belief in the wrong goal (Scenario 4).
\acs{GOFI} infers a higher posterior than GR-Only because it models the possibility that the non-ego's actions are due to the presence of an occluded factor. It also never converges to belief in the incorrect goal.

Our scenarios only consider the case where an occluded factor is present.
In Appendix \ref{app:variations} we include evaluation in Scenarios 2 and 4 in which the occluded factor is \textit{not} present and the non-ego's goal is switched to G2.
In these settings we expect to see that \acs{GOFI} converges to the prior belief on the presence of occluded factors and these additional results confirm this hypothesis.

\subsection{MCTS Integration Evaluation}\label{sec:empirical:mcts}

Next, we evaluate the integration of \acs{GOFI} with \textsc{mcts}.
\begin{figure}
\centering
    \begin{subfigure}{0.47\linewidth}
    \includegraphics[width=\linewidth,clip=true,trim=2mm 0cm 1cm 0cm]{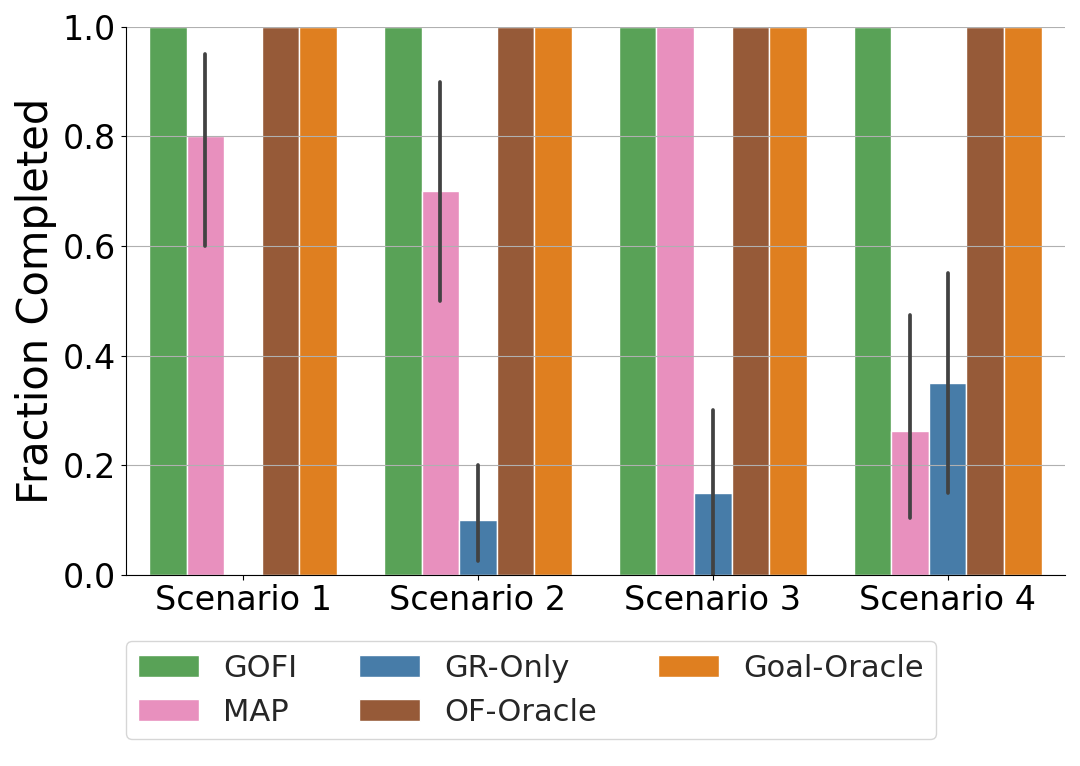}
    \caption{All Scenarios}
    \label{fig:collisions:all}
    \end{subfigure}
    \begin{subfigure}{0.47\linewidth}
    \includegraphics[width=\linewidth,clip=true,trim=2mm 0cm 1cm 0cm]{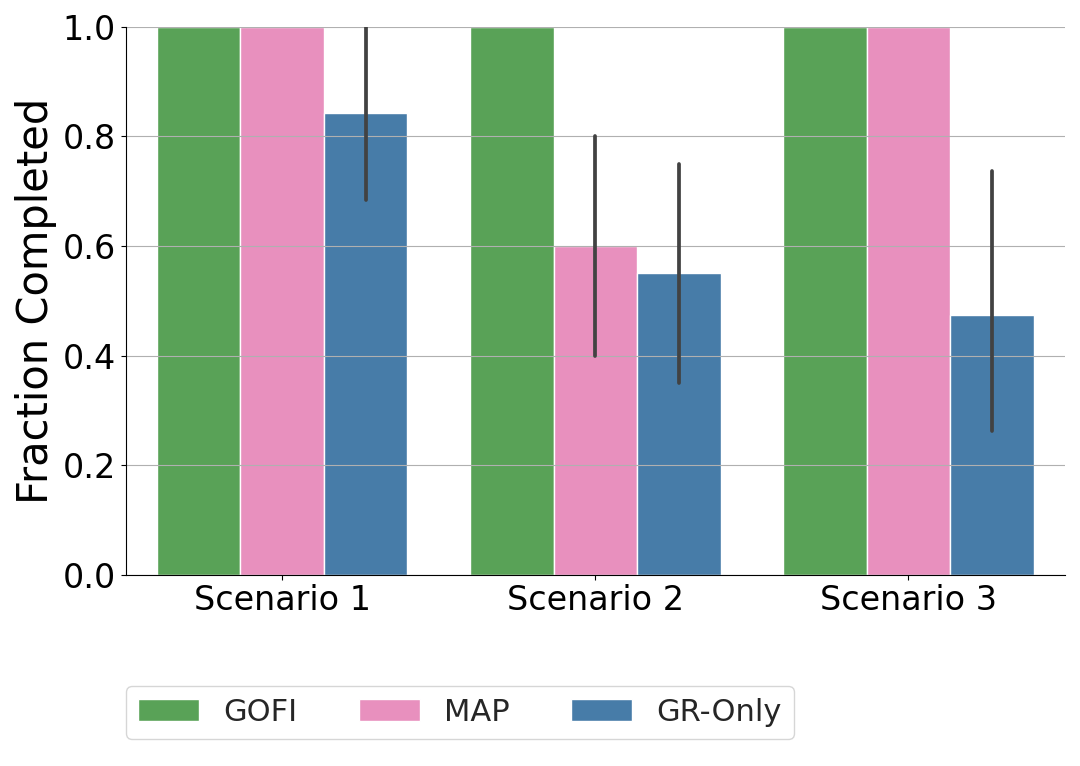}
    \caption{Scenarios with Perception}
    \label{fig:collisions:perception}
    \end{subfigure}
    \caption{Fraction of trials without a collision. Error bars give a bootstrap 95\% confidence interval.
    }
    \label{fig:collisions}
\end{figure}
\Cref{fig:collisions} shows the fraction of runs completed in each scenario without a collision.
Of the non-oracle methods (\acs{GOFI}, \acs{MAP}, GR-Only), \acs{GOFI} obtains the highest completion percentage in all scenarios.
OF-Oracle knows an occluded factor exists and Goal-Oracle knows the true goal in all scenarios. 
This privileged knowledge allows these methods to upper bound the performance of \acs{GOFI} but could not be realised in practice.

In terms of baselines without privileged information, the comparison to \acs{MAP} shows the importance of our sampling-based integration of an inferred belief with \acs{MCTS}.
Since \acs{MAP} plans with the most likely goal and occluded factor under the posterior, it ignores low-probability -- but potentially costly -- situations when selecting actions.
GR-Only also experiences more collisions than \acs{GOFI} as it cannot infer the presence of occluded vehicles.
This comparison shows how using the behaviour of other vehicles to ``sense" occluded factors yields safer planning.

Finally, while the completion rates shown in \cref{fig:collisions} are measured \textit{without} perception of occluded factors, we also evaluate versions of Scenarios 1, 2, and 3 in which the ego-vehicle can perceive the occluded factor.
In Scenario 1, the pedestrian is occluded if a line from the ego-vehicle to the pedestrian intersects a box around the non-ego vehicle.
In Scenario 2, we base occlusion on the ego-vehicles position; once the ego-vehicle reaches the position where the non-ego vehicle began its overtake maneuver, we set the ego's belief in the presence of the occluded vehicle to 1.0.
In Scenario 3, we determine occlusion based on whether the ego-vehicle's line of sight to the occluded vehicle is obscured by a building.
This change improves the completion percentage of GR-Only and \textsc{map} (see \cref{fig:collisions:perception}).
However, due to the fixed 1 Hz control frequency of the \acs{MCTS} planner, these baselines still cannot always avoid collisions.
\acs{GOFI} complements perception with inference based on the non-ego's behaviour and so is able to plan for the potential of an occluded factor sooner and always avoid collisions.
While \textsc{map} also avoids all collisions in Scenarios 1 and 3, it fails to do so in Scenario 2.

\section{Conclusion}

In this paper we considered the problem of inferring the goals of an observed vehicle in the presence of occluded factors which may confound goal recognition.
This problem arises in autonomous driving scenarios with occluded vehicles and pedestrians.
We showed how unobserved occluded factors may confound goal recognition, introduced an interpretable inverse-planning algorithm for joint goal and occluded factor inference, and demonstrated how it can be integrated into \textsc{mcts} action selection.
In addition to more accurate goal recognition, our algorithm infers the presence of unobserved occluded factors based on the behaviour of observed vehicles.
We performed an empirical evaluation of our framework in simulated autonomous driving scenarios that require occluded factor inference for safe driving.
Our empirical results demonstrated the necessity of joint goal and occluded factor inference as goal recognition alone misidentified goals in the presence of occluded factors and resulted in more collisions in safety critical scenarios.
Our Goal and Occluded Factor Inference algorithm can simultaneously infer goals and occluded factors, leading to fewer collisions in safety critical scenarios. 

In the future, we aim to address the construction of the set of possible occluded factors.
This problem requires identifying potential locations of occluded factors (i.e., occluded locations) as well as the set of behaviours consistent with identified locations.
We will also consider how to select information gathering actions to resolve uncertainty about occluded factors.

\section*{Acknowledgment}
This research was in part supported by the Royal Society via an Industry Fellowship (S.A.), the Alan Turing Institute via project "Model Criticism in Multi-Agent Systems" (J.H.), and UK Research and Innovation via NPIF Innovation Placements (E.F.). J.H. was with Five AI and School of Informatics, University of Edinburgh while this research took place. A.R. and E.F. were interns at Five AI while this research took place. We would like to thank Majd Hawasly and others at Five AI for insightful comments on this work.

\appendices

\section{Extended MCTS Integration Description}\label{appendix:mcts}

In this appendix we provide full pseudocode (\cref{alg:mcts}) for our integration of the beliefs obtained with \acs{GOFI} into \textsc{mcts}.
\textsc{mcts} runs a fixed number of iterations to build up a search tree in which nodes correspond to sequences of actions.
Recall from Section \ref{sec:planning} that we follow Albrecht et al.\ \cite{albrecht2020integrating} and plan using an action-space of \textit{macro-actions} where a macro-action is a sequence of pre-defined maneuvers such as turn, change-lane, or follow-lane.
Maneuvers define trajectories consisting of a reference path and target speeds along the path, which are mapped to steering and acceleration controls via \textsc{pid} control or adaptive cruise control.
Thus, even though planning is done using a high-level action-space, the \textsc{mcts} iterations use low-level simulation of actions.

In the search-tree, each node, $q$, contains a value, $Q(q,\mu)$, for each macro-action $\mu$ that gives the expected future reward, following choosing $\mu$ in node $q$.
Once these \textit{action-values} are estimated, the optimal macro-action at time $t$ can be selected as $\argmax_\mu Q(\text{root}, \mu)$.
Each iteration begins by sampling an instantiation of the occluded factors, non-ego goals, and trajectories for non-ego vehicles.
The ego-vehicle then uses a given exploration strategy (we use \textsc{ucb} \cite{ucb2002}) to select macro-action until a fixed search depth is reached or a termination condition is met.
Upon termination, the final reward $r$ is back-propagated through the search tree using the following update rule based on Q-learning \cite{watkins1992q}:

{
\small
\begin{equation}\label{eqn:backprop}
    Q(q,\mu) \leftarrow Q(q,\mu) + 
    \begin{cases}
        \delta^{-1}\qty[r-Q(q,\mu)] \text{ if } $q$ \text{ leaf node, else} \\
        \delta^{-1}\qty[\max_{\mu'}Q(q',\mu')-Q(q,\mu)]
    \end{cases}
\end{equation}
}
where $\delta$ is the number of times maneuver $\mu$ has been selected in node $q$.

\begin{algorithm}[h!]
	\textbf{Returns:} optimal maneuver for the ego-vehicle $\ego$ in state $\est{t}{}{}$
	
	\vspace{3pt}
	
	Perform $k$ iterations:
	\begin{algorithmic}[1]
		\STATE Search node $q.\st{}{} \gets \est{}{}{}{t}$ (root node)
		\STATE Search depth $d \gets 0$
	    \STATE Sample occluded factor instantiation $\hs{}{} \sim \Pr(\hs{}{} | \est{}{}{1:t})$
		\FORALL{$i \in \awareset \setminus \{ \ego \}$}
		    \STATE Sample goal $\go{i}{} \sim p(\go{i}{} \,|\, \est{}{}{1:t}, \hs{}{})$
		    \STATE $\st{i}{1:T} \leftarrow \textsc{PlanOptimal}(\est{}{i}{t}, g^i, z)$ %
		\ENDFOR
		\WHILE{$d < d_{\max}$}
		    \STATE Select maneuver $\mu$ for $\ego$ applicable in $q.\st{}{}$ \label{alg:mcts-1}
		    \STATE $\st{}{\tau,\iota} \gets$ Simulate maneuver until it terminates, with other vehicles following their sampled trajectories $\st{i}{1:T}$
		    \STATE $r \gets \emptyset$
		    \IF{ego-vehicle collides during $\st{}{\tau,\iota}$}
		        \STATE $r \gets c_{\text{coll.}}$
		    \ELSIF{$\st{\ego}{\iota}$ achieves the ego-vehicle's goal $g^\ego$}
		        \STATE $r \gets c(\st{}{t:T})$
		    \ELSIF{$d = d_{\max}-1$}
		        \STATE $r \gets r_{\text{term.}}$
		    \ENDIF
		    \IF{$r \neq \emptyset$}
		        \STATE Use \eqref{eqn:backprop} to backprop $r$ along search branches $(q,\mu,q')$ traversed in the iteration
		        \STATE Start next iteration
		    \ENDIF
		    \STATE $q'.s = \st{}{\iota}$; $q \gets q'$; $d \gets d + 1$
		\ENDWHILE
	\end{algorithmic}
	Return maneuver $\mu \in \argmin_\mu Q(\text{root},\mu)$
	\caption{Monte Carlo Tree Search algorithm}
	\label{alg:mcts}
\end{algorithm}

\section{Additional Empirical Results}\label{app:variations}

In the main paper, we evaluated \acs{GOFI} and baselines under a single instantiation of goals and occluded factors in each scenario.
In this appendix we include results for Scenarios 2 and 4 with no occluded factor present and the visible non-ego's goal switched from G1 to G2.
With these changes, GR-Oracle and OF-Oracle, that assume G1 and occluded factor presence respectively, use incorrect assumptions.

\begin{figure}[h]
    \centering
    \begin{subfigure}{0.23\textwidth}
        \includegraphics[width=\textwidth,clip=true,trim=2cm 1cm 3cm 1cm]{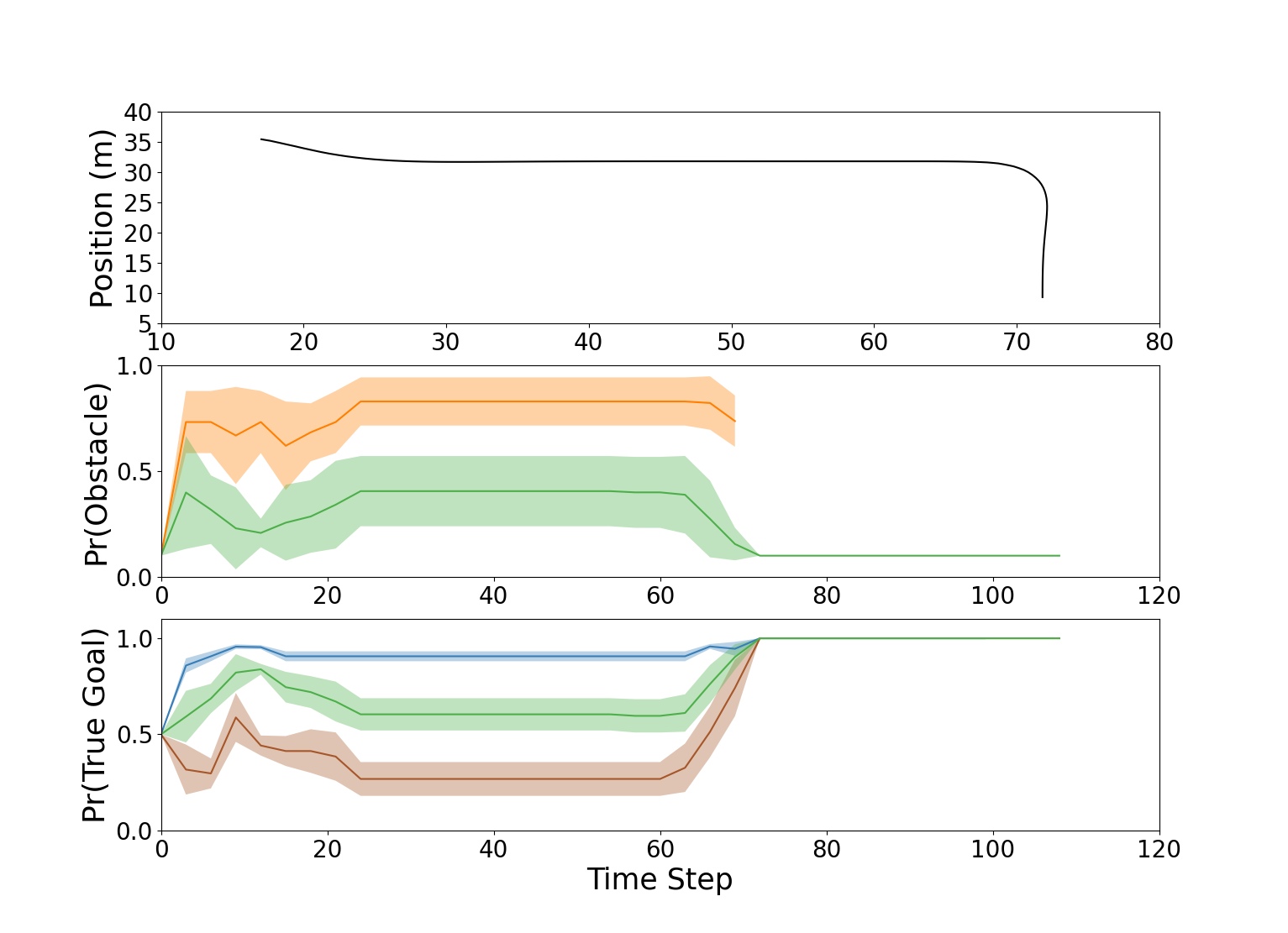}
        \caption{Scenario 2}
    \end{subfigure}
    \begin{subfigure}{0.23\textwidth}
        \includegraphics[width=\textwidth,clip=true,trim=2cm 1cm 3cm 1cm]{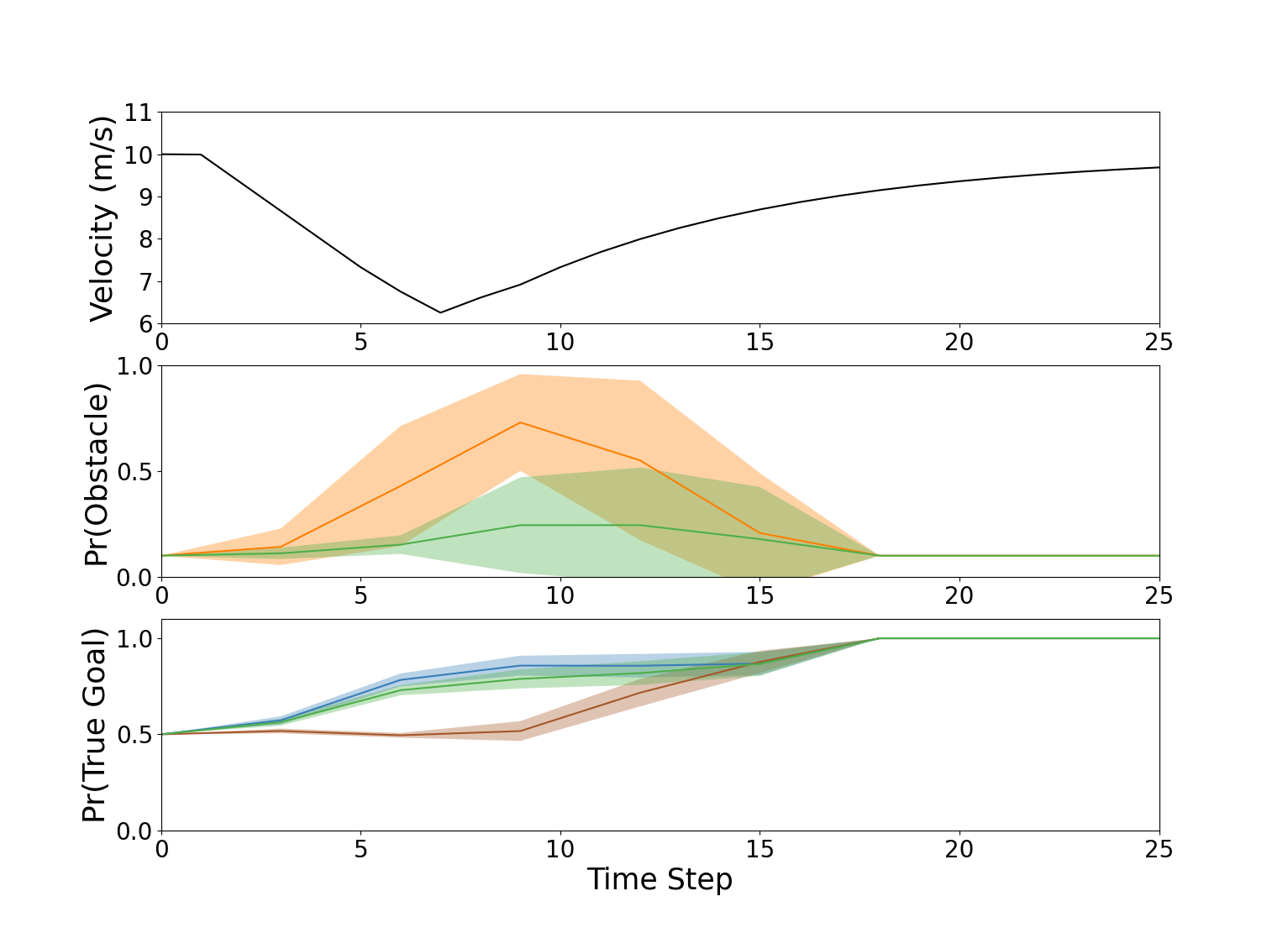}
        \caption{Scenario 4}
    \end{subfigure}
    \begin{subfigure}{\linewidth}
        \includegraphics[width=\linewidth]{figs/iros-figs/gofi-legend.png}
    \end{subfigure}
    \caption{\textbf{Scenario Variation Results.}
        \textit{Top}: Trajectory of the observed non-ego vehicle.
        \textit{Middle}: Belief in the presence of an occluded obstacle over time (higher is better).
        \textit{Bottom}: Belief in the true follow-road goal over time.
        Beliefs averaged over 20 repetitions (shaded area represents standard error).}
    \label{fig:vary}
\end{figure}

\cref{fig:vary} shows the inferred probabilities for each variation.
For occluded factor recognition, we see in both scenarios that Goal-Oracle is misled by its false assumption that the non-ego is heading to G1.
The false assumption leads it to explain the non-ego's behaviour as indicative of the presence of an occluded factor (note that we end the plotted beliefs at the point it is impossible for the non-ego to reach G1).
In Scenario 2, \acs{GOFI} initially increases its belief in the presence of an occluded factor after the lane-change. However, as the non-ego continues in the right lane, the belief returns to the prior. Once the non-ego turns, the belief returns to the prior since the non-ego's goal is sufficient for explaining the non-ego's observed behaviour.
A similar trend is observed in Scenario 4 in response to the non-ego's slow-down and then eventual acceleration as it turns left.
Note that, \acs{GOFI} should \textit{not} converge to a belief in the absence of an occluded factor given the observed non-ego behaviour as absence of evidence is not evidence of absence.

For goal recognition, since there is no occluded factor present, occluded factors do not have to be considered for identifying the non-ego's goal.
Thus GR-Only has the most accurate inference for goal recognition.
Since OF-Oracle (falsely) assumes the presence of an occluded factor, it explains the lane change as obstacle-avoidance and thus does not increase its belief in the true goal.
In both scenarios, \acs{GOFI} obtains a posterior belief that is between OF-Oracle and GR-Only.
Recall from \cref{fig:probs} in the main text that the relative performance of the baselines is flipped under a different setting of the non-ego goals and occluded factor presence.
Taken together these results show that by jointly recognising the factors explaining the non-ego behaviours \acs{GOFI} can converge to higher posteriors on the true factors.

\typeout{}
\bibliographystyle{plain}
\bibliography{prediction}

\begin{thebibliography}{10}

\bibitem{afolabi2018people}
Oladapo Afolabi, Katherine Driggs-Campbell, Roy Dong, Mykel~J. Kochenderfer,
  and S.~Shankar Sastry.
\newblock People as sensors: Imputing maps from human actions.
\newblock In {\em 2018 IEEE/RSJ International Conference on Intelligent Robots
  and Systems (IROS)}, pages 2342--2348. IEEE, 2018.

\bibitem{albrecht2020integrating}
Stefano~V. Albrecht, Cillian Brewitt, John Wilhelm, Balint Gyevnar, Francisco
  Eiras, Mihai Dobre, and Subramanian Ramamoorthy.
\newblock Interpretable goal-based prediction and planning for autonomous
  driving.
\newblock In {\em Proceedings of the IEEE Conference on Robotics and Automation
  (ICRA)}, 2021.

\bibitem{albrecht2016causality}
Stefano~V. Albrecht and Subramanian Ramamoorthy.
\newblock Exploiting causality for selective belief filtering in dynamic
  {B}ayesian networks.
\newblock {\em Journal of Artificial Intelligence Research}, 55:1135--1178,
  2016.

\bibitem{albrecht2018modelling}
Stefano~V. Albrecht and Peter Stone.
\newblock Autonomous agents modelling other agents: A comprehensive survey and
  open problems.
\newblock {\em Artificial Intelligence}, 258:66--95, 2018.
\newblock DOI: 10.1016/j.artint.2018.01.002.

\bibitem{ucb2002}
Peter Auer, Nicolo Cesa-Bianchi, and Paul Fischer.
\newblock Finite-time analysis of the multiarmed bandit problem.
\newblock {\em Machine Learning}, 47(2-3):235--256, 2002.

\bibitem{baker2009action}
Chris~L. Baker, Rebecca Saxe, and Joshua~B. Tenenbaum.
\newblock Action understanding as inverse planning.
\newblock {\em Cognition}, 113(3):329--349, 2009.

\bibitem{bouton2018scalable}
Maxime Bouton, Alireza Nakhaei, Kikuo Fujimura, and Mykel~J. Kochenderfer.
\newblock Scalable decision making with sensor occlusions for autonomous
  driving.
\newblock In {\em 2018 IEEE International Conference on Robotics and Automation
  (ICRA)}, pages 2076--2081. IEEE, 2018.

\bibitem{casas_intentnet_2018}
Sergio Casas, Wenjie Luo, and Raquel Urtasun.
\newblock \{{IntentNet}:\} {Learning} to {Predict} {Intention} from {Raw}
  {Sensor} {Data}.
\newblock In {\em Conference on {Robot} {Learning}}, pages 947--956, 2018.

\bibitem{chai_multipath_2019}
Yuning Chai, Benjamin Sappm, Mayank Bansal, and Dragomir Anguelov.
\newblock {MultiPath}: {Multiple} {Probabilistic} {Anchor} {Trajectory}
  {Hypotheses} for {Behavior} {Prediction}.
\newblock In {\em Proceedings of the 3rd {Conference} on {Robot} {Learning}},
  2019.

\bibitem{deoConvolutionalSocialPooling2018}
Nachiket Deo and Mohan~M. Trivedi.
\newblock Convolutional social pooling for vehicle trajectory prediction.
\newblock In {\em 2018 {{IEEE}}/{{CVF Conference}} on {{Computer Vision}} and
  {{Pattern Recognition Workshops}} ({{CVPRW}})}, pages 1549--1548, {Salt Lake
  City, Utah, USA}, June 2018. {IEEE}.

\bibitem{filosCanAutonomousVehicles2020}
Angelos Filos, Panagiotis Tigkas, Rowan Mcallister, Nicholas Rhinehart, Sergey
  Levine, and Yarin Gal.
\newblock Can autonomous vehicles identify, recover from, and adapt to
  distribution shifts?
\newblock In {\em International {{Conference}} on {{Machine Learning}}}, volume
  119, pages 3145--3153. {PMLR}, November 2020.

\bibitem{galceran2015augmented}
Enric Galceran, Edwin Olson, and Ryan~M. Eustice.
\newblock Augmented vehicle tracking under occlusions for decision-making in
  autonomous driving.
\newblock In {\em 2015 IEEE/RSJ International Conference on Intelligent Robots
  and Systems (IROS)}, pages 3559--3565. IEEE, 2015.

\bibitem{gonzalezdebada2020oclussion}
Ezequiel Gonz\'{a}lez~Debada, Adeline Ung, and Denis Gillet.
\newblock Occlusion-aware motion planning at roundabouts.
\newblock {\em IEEE Transactions on Intelligent Vehicles}, 2020.

\bibitem{kimProbabilisticVehicleTrajectory2017}
ByeoungDo Kim, Chang~Mook Kang, Jaekyum~Kim Kim, Seung~Hi Lee, Chung~Choo
  Chung, and Jun~Won Choi.
\newblock Probabilistic vehicle trajectory prediction over occupancy grid map
  via recurrent neural network.
\newblock In {\em 2017 {{IEEE}} 20th {{International Conference}} on
  {{Intelligent Transportation Systems}} ({{ITSC}})}, pages 399--404,
  {Yokohama, Japan}, October 2017. {IEEE}.

\bibitem{ks2006}
Levente Kocsis and Csaba Szepesv{\'a}ri.
\newblock Bandit based {Monte-Carlo} planning.
\newblock In {\em Proceedings of the 17th European Conference on Machine
  Learning}, pages 282--293. Springer, 2006.

\bibitem{koopman2019certification}
Philip Koopman, Rob Hierons, Siddartha Khastgir, John Clark, Michael Fisher,
  Rob Alexander, Kerstin Eder, Pete Thomas, Geoff Barrett, Philip Torr, et~al.
\newblock Certification of highly automated vehicles for use on uk roads:
  Creating an industry-wide framework for safety.
\newblock Technical report, Five AI Ltd, 2019.

\bibitem{kurzer2020parallelization}
Karl Kurzer, Christoph H{\"o}rtnagl, and J~Marius Z{\"o}llner.
\newblock Parallelization of monte carlo tree search in continuous domains.
\newblock {\em arXiv preprint arXiv:2003.13741}, 2020.

\bibitem{lee2020monte}
Jongmin Lee, Wonseok Jeon, Geon-Hyeong Kim, and Kee-Eung Kim.
\newblock Monte-carlo tree search in continuous action spaces with value
  gradients.
\newblock In {\em Proceedings of the AAAI Conference on Artificial
  Intelligence}, volume~34, pages 4561--4568, 2020.

\bibitem{lee2017desire}
Namhoon Lee, Wongun Choi, Paul Vernaza, Christopher~B. Choy, Philip H.~S. Torr,
  and Manmohan Chandraker.
\newblock {DESIRE}: {Distant} future prediction in dynamic scenes with
  interacting agents.
\newblock In {\em Proceedings of the {IEEE} {Conference} on {Computer} {Vision}
  and {Pattern} {Recognition}}, pages 336--345, 2017.

\bibitem{li2018generic}
Jiachen Li, Wei Zhan, and Masayoshi Tomizuka.
\newblock Generic vehicle tracking framework capable of handling occlusions
  based on modified mixture particle filter.
\newblock In {\em 2018 IEEE Intelligent Vehicles Symposium (IV)}, pages
  936--942. IEEE, 2018.

\bibitem{morales2018towards}
Luis~Yoichi Morales, Akai Naoki, Yuki Yoshihara, and Hiroshi Murase.
\newblock Towards predictive driving through blind intersections.
\newblock In {\em 2018 21st International Conference on Intelligent
  Transportation Systems (ITSC)}, pages 716--722. IEEE, 2018.

\bibitem{pulverPILOTEfficientPlanning2020}
Henry Pulver, Francisco Eiras, Ludovico Carozza, Majd Hawasly, Stefano~V.
  Albrecht, and Subramanian Ramamoorthy.
\newblock {{PILOT}}: {{Efficient}} planning by imitation learning and
  optimisation for safe autonomous driving.
\newblock {\em arXiv:2011.00509 [cs]}, November 2020.

\bibitem{ramirez2009plan}
Miquel Ram{\'\i}rez and Hector Geffner.
\newblock Plan recognition as planning.
\newblock In {\em Twenty-First International Joint Conference on Artificial
  Intelligence}, 2009.

\bibitem{ramirez2010probabilistic}
Miquel Ram\'{i}rez and Hector Geffner.
\newblock Probabilistic plan recognition using off-the-shelf classical
  planners.
\newblock In {\em 24th AAAI Conference on Artificial Intelligence}, pages
  1121--1126, 2010.

\bibitem{ramirez2011goal}
Miquel Ram{\'\i}rez and Hector Geffner.
\newblock Goal recognition over pomdps: Inferring the intention of a pomdp
  agent.
\newblock In {\em IJCAI}, pages 2009--2014. Citeseer, 2011.

\bibitem{rhinehart2019precog}
Nicholas Rhinehart, Rowan McAllister, Kris Kitani, and Sergey Levine.
\newblock {PRECOG}: {Prediction} conditioned on goals in visual multi-agent
  settings.
\newblock In {\em Proceedings of the {IEEE} {International} {Conference} on
  {Computer} {Vision}}, pages 2821--2830, 2019.

\bibitem{sohrabi2016plan}
Shirin Sohrabi, Anton~V. Riabov, and Octavian Udrea.
\newblock Plan recognition as planning revisited.
\newblock In {\em IJCAI}, pages 3258--3264, 2016.

\bibitem{sukthankar2014plan}
Gita Sukthankar, Christopher Geib, Hung~Hai Bui, David Pynadath, and Robert~P
  Goldman.
\newblock {\em Plan, activity, and intent recognition: Theory and practice}.
\newblock Newnes, 2014.

\bibitem{sun2019behavior}
Liting Sun, Wei Zhan, Ching-Yao Chan, and Masayoshi Tomizuka.
\newblock Behavior planning of autonomous cars with social perception.
\newblock In {\em 2019 IEEE Intelligent Vehicles Symposium (IV)}, pages
  207--213. IEEE, 2019.

\bibitem{tas2020tackling}
Omer~Sahin Tas and Christoph Stiller.
\newblock Tackling existence probabilities of objects with motion planning for
  automated urban driving.
\newblock {\em arXiv preprint arXiv:2002.01254}, 2020.

\bibitem{thiebes2020trustworthy}
Scott Thiebes, Sebastian Lins, and Ali Sunyaev.
\newblock Trustworthy artificial intelligence.
\newblock {\em Electronic Markets}, October 2020.

\bibitem{vered2017heuristic}
Mor Vered and Gal~A. Kaminka.
\newblock Heuristic online goal recognition in continuous domains.
\newblock {\em arXiv preprint arXiv:1709.09839}, 2017.

\bibitem{watkins1992q}
Christopher J. C.~H. Watkins and Peter Dayan.
\newblock Q-learning.
\newblock {\em Machine learning}, 8(3-4):279--292, 1992.

\bibitem{whitehouse2011determinization}
Daniel Whitehouse, Edward~J Powley, and Peter~I Cowling.
\newblock Determinization and information set monte carlo tree search for the
  card game dou di zhu.
\newblock In {\em 2011 IEEE Conference on Computational Intelligence and Games
  (CIG'11)}, pages 87--94. IEEE, 2011.

\bibitem{zhang2021safe}
Zixu Zhang and Jaime~F Fisac.
\newblock Safe occlusion-aware autonomous driving via game-theoretic active
  perception.
\newblock In {\em In proceedings of Robotics: Science and Systems (RSS)}, 2021.

\end{thebibliography}

\end{document}